\documentclass[final]{cvpr}

\usepackage{times}
\usepackage{epsfig}
\usepackage{graphicx}
\usepackage{amsmath}
\usepackage{amssymb}
\usepackage{mathtools}  
\mathtoolsset{showonlyrefs}  

\usepackage[]{caption}
\usepackage{multirow}
\usepackage{mathtools}
\usepackage{color}
\usepackage{soul}
\usepackage{subcaption}
\usepackage{algorithmicx}
\usepackage{nicefrac,xfrac}
\usepackage{algorithm,algpseudocode}
\usepackage{bbm}
\usepackage{array}
\usepackage{booktabs}
\usepackage{enumitem}

\newcommand{\cutsectionup}{\vspace*{-0.05in}}
\newcommand{\cutsectiondown}{\vspace*{-0.05in}}

\newcommand{\cutsubsectionup}{\vspace*{-0.05in}}
\newcommand{\cutsubsectiondown}{\vspace*{-0.05in}}

\newcommand{\cutparagraphup}{\vspace*{-0.12in}}

\usepackage[pagebackref=true,breaklinks=true,colorlinks,bookmarks=false]{hyperref}

\newcommand{\CLGAN}{XMC-GAN\xspace}  
\newcommand{\clgan}{\CLGAN}
\newcommand{\oi}{Open~Images\xspace}
\newcommand{\coco}{MS-COCO\xspace}
\newcommand{\cocoold}{COCO-14\xspace}
\newcommand{\coconew}{COCO-17\xspace}
\newcommand{\cocoln}{LN-COCO\xspace}
\newcommand{\lnoi}{LN-OpenImages\xspace}
\newcommand{\IS}{IS\xspace}
\newcommand{\FID}{FID\xspace}

\newcommand{\rprec}{R-prec\xspace}

\newcommand{\ssent}{\mathcal{S}_{\text{sent}}}
\newcommand{\simg}{\mathcal{S}_{\text{img}}}
\newcommand{\sword}{\mathcal{S}_{\text{word}}}
\newcommand{\lsent}{\mathcal{L}_{\text{sent}}}
\newcommand{\limg}{\mathcal{L}_{\text{img}}}
\newcommand{\lword}{\mathcal{L}_{\text{word}}}
\newcommand{\el}{\mathcal{L}}

\newcommand{\pz}{\hphantom{0}}

\newcommand{\myindent}[1]{
\newline\makebox[#1cm]{}
}
\newcommand{\compareexamplesmall}[2]{
\scriptsize{#1} & 
\raisebox{-.5\height}{\includegraphics[width=0.09\textwidth,height=0.09\textwidth]{images/coco_images/opgan/#2_s-1.png}}
& 
\raisebox{-.5\height}{\includegraphics[width=0.09\textwidth,height=0.09\textwidth]{images/coco_images/sdgan/#2_s-1.png}} &
\raisebox{-.5\height}{\includegraphics[width=0.09\textwidth,height=0.09\textwidth]{images/coco_images/cpgan/#2_s-1.png}} &
\raisebox{-.5\height}{\includegraphics[width=0.09\textwidth,height=0.09\textwidth]{images/coco_images/ours/#2_s-1.png}} 
}

\newcommand{\compareexamplenoise}[2]{
\scriptsize{#1} & 
\raisebox{-.5\height}{\includegraphics[width=0.14\textwidth,height=0.14\textwidth]{images/noise/cpgan/#2_s-1_0.png}} & 
\raisebox{-.5\height}{\includegraphics[width=0.14\textwidth,height=0.14\textwidth]{images/noise/cpgan/#2_s-1_1.png}} & 
\raisebox{-.5\height}{\includegraphics[width=0.14\textwidth,height=0.14\textwidth]{images/noise/cpgan/#2_s-1_2.png}} &
\raisebox{-.5\height}{\includegraphics[width=0.14\textwidth,height=0.14\textwidth]{images/noise/ours/#2_s-1_0.png}} & 
\raisebox{-.5\height}{\includegraphics[width=0.14\textwidth,height=0.14\textwidth]{images/noise/ours/#2_s-1_1.png}} & 
\raisebox{-.5\height}{\includegraphics[width=0.14\textwidth,height=0.14\textwidth]{images/noise/ours/#2_s-1_2.png}}
}

\newcommand{\compareexamplelnnooriginal}[3]{
\scriptsize{#1} & 
\raisebox{-.5\height}{\includegraphics[width=0.09\textwidth,height=0.09\textwidth]{images/coco_ln_images/attngan/#2.png}} &
\raisebox{-.5\height}{\includegraphics[width=0.09\textwidth,height=0.09\textwidth]{images/coco_ln_images/trecs/#2.png}} &
\raisebox{-.5\height}{\includegraphics[width=0.09\textwidth,height=0.09\textwidth]{images/coco_ln_images/ours/#2.png}}
}

\newcommand{\compareexampleln}[3]{
\scriptsize{#1} & 
\raisebox{-.5\height}{\includegraphics[width=0.12\textwidth,height=0.12\textwidth]{images/coco_ln_images/original/#3.jpg}} &
\raisebox{-.5\height}{\includegraphics[width=0.12\textwidth,height=0.12\textwidth]{images/coco_ln_images/attngan/#2.png}} &
\raisebox{-.5\height}{\includegraphics[width=0.12\textwidth,height=0.12\textwidth]{images/coco_ln_images/trecs/#2.png}} &
\raisebox{-.5\height}{\includegraphics[width=0.12\textwidth,height=0.12\textwidth]{images/coco_ln_images/ours/#2.png}}
}

\newcommand{\compareexamplelnoi}[2]{
\scriptsize{#1} & 
\raisebox{-.5\height}{\includegraphics[width=0.12\textwidth,height=0.12\textwidth]{images/ln_oi_images/real/#2.jpg}} &
\raisebox{-.5\height}{\includegraphics[width=0.12\textwidth,height=0.12\textwidth]{images/ln_oi_images/ours/#2.png}}
}

\newcommand{\compareexamplecaptions}[8]{
 & \scriptsize{#1} & \scriptsize{#2} & \scriptsize{#3} & \scriptsize{#4} & \scriptsize{#5} & \tiny{#6} \\
\raisebox{-.5\height}{\includegraphics[width=0.13\textwidth,height=0.13\textwidth]{images/ours_coco_captions/#7.jpg}} & 
\raisebox{-.5\height}{\includegraphics[width=0.13\textwidth,height=0.13\textwidth]{images/ours_coco_captions/#7_s-1.png}} & 
\raisebox{-.5\height}{\includegraphics[width=0.13\textwidth,height=0.13\textwidth]{images/ours_coco_captions/#7_s-2.png}} & 
\raisebox{-.5\height}{\includegraphics[width=0.13\textwidth,height=0.13\textwidth]{images/ours_coco_captions/#7_s-3.png}} & 
\raisebox{-.5\height}{\includegraphics[width=0.13\textwidth,height=0.13\textwidth]{images/ours_coco_captions/#7_s-4.png}} & 
\raisebox{-.5\height}{\includegraphics[width=0.13\textwidth,height=0.13\textwidth]{images/ours_coco_captions/#7_s-5.png}} & 
\raisebox{-.5\height}{\includegraphics[width=0.13\textwidth,height=0.13\textwidth]{images/ours_coco_captions/#8.png}} \\
}

\def\cvprPaperID{4981} 
\def\confYear{CVPR 2021}
\newcommand*\samethanks[1][\value{footnote}]{\footnotemark[#1]}

\def\And{
  \end{tabular}\\%
  \begin{tabular}[t]{c}}

\begin{document}

\title{Cross-Modal Contrastive Learning for Text-to-Image Generation}


\author{
Han Zhang\thanks{Equal contribution.}\\
Google Research\\
{\tt\small zhanghan@google.com}
\and
Jing Yu Koh\samethanks\;\thanks{Work done as a member of the Google AI Residency program.}\\
Google Research\\
{\tt\small jykoh@google.com}
\And
Jason Baldridge\\
Google Research\\
{\tt\small jridge@google.com}
\and
Honglak Lee\thanks{Work performed at Google Research.}\\
University of Michigan\\
{\tt\small honglak@umich.edu}
\and
Yinfei Yang\\
Google Research\\
{\tt\small yinfeiy@google.com}
}

\maketitle

\begin{abstract}
\vspace*{-0.1in}
The output of text-to-image synthesis systems should be coherent, clear, photo-realistic scenes with high semantic fidelity to their conditioned text descriptions. Our Cross-Modal Contrastive Generative Adversarial Network (\CLGAN) addresses this challenge by maximizing the mutual information between image and text. It does this via multiple contrastive losses which capture inter-modality and intra-modality correspondences. \CLGAN uses an attentional self-modulation generator, which enforces strong text-image correspondence, and a contrastive discriminator, which acts as a critic as well as a feature encoder for contrastive learning. The quality of \CLGAN's output is a major step up from previous models, as we show on three challenging datasets. On MS-COCO, not only does \CLGAN improve state-of-the-art FID from 24.70 to 9.33, but--more importantly--people prefer \CLGAN by 77.3\% for image quality and 74.1\% for image-text alignment, compared to three other recent models. \CLGAN also generalizes to the challenging Localized Narratives dataset (which has longer, more detailed descriptions), improving state-of-the-art FID from 48.70 to 14.12. Lastly, we train and evaluate \CLGAN on the challenging Open Images data, establishing a strong benchmark FID score of 26.91.
\vspace*{-0.1in}
\end{abstract}


\cutsectionup
\section{Introduction} \label{sec:intro}
\cutsectiondown

Compared to other kinds of inputs (\eg, sketches and object masks), descriptive sentences are an intuitive and flexible way to express visual concepts for generating images. The main challenge for text-to-image synthesis lies in learning from unstructured description and handling the different statistical properties between vision and language inputs.

\begin{figure}
    \centering
    \includegraphics[width=\linewidth]{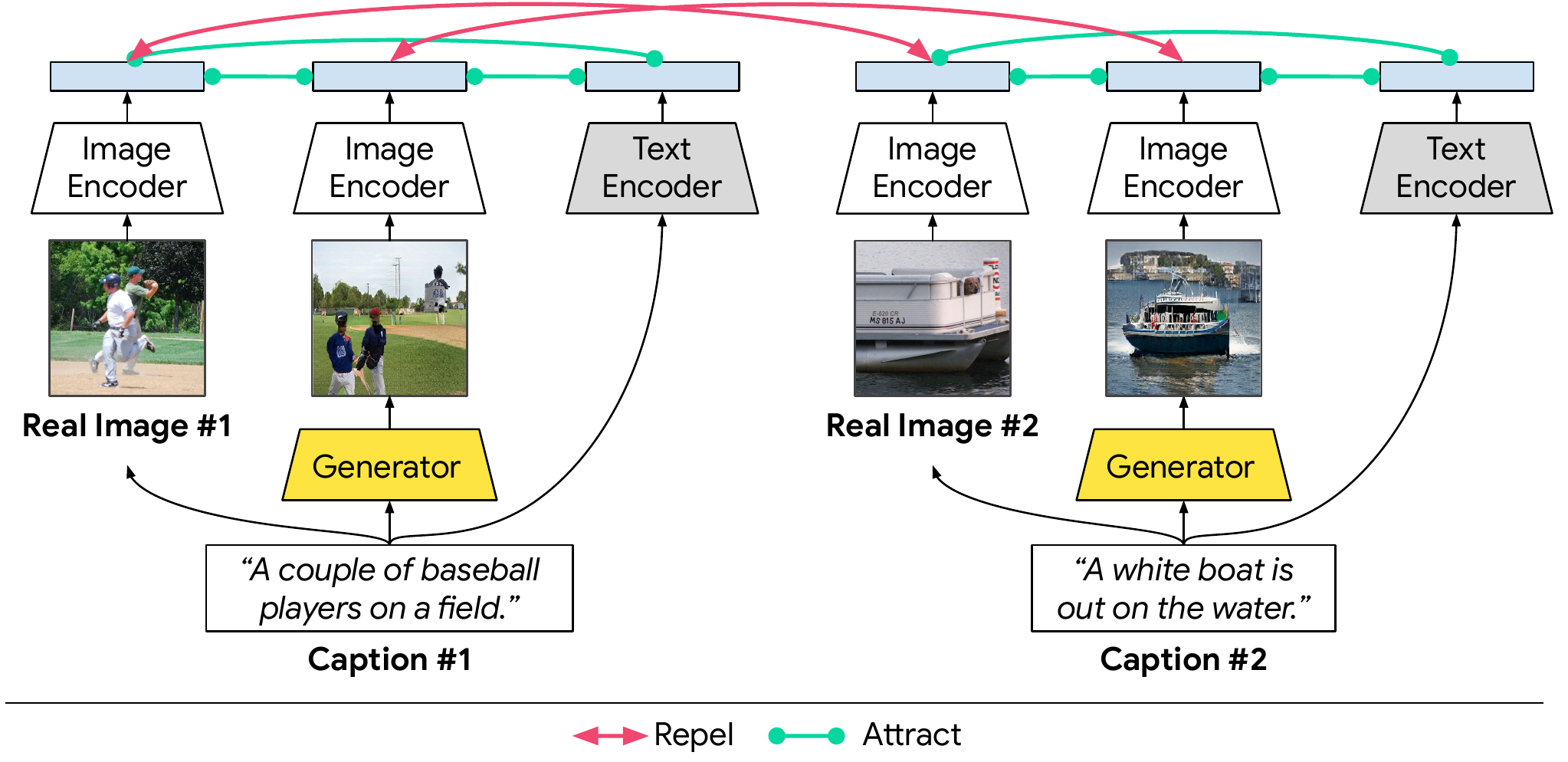}
    \vspace{-10pt}
    \caption{Inter-modal and intra-modal contrastive losses in our proposed \CLGAN text-to-image synthesis model.}
    \label{fig:overview}
\end{figure}

Generative Adversarial Networks (GANs)~\cite{goodfellow2014generative} have shown promising results on text-to-image generation \cite{reed2016generative,Han17,Han17stackgan2}, using a conditional GAN formulation~\cite{gauthier2015conditional}. 
AttnGAN~\cite{xu18} proposes a multi-stage refinement framework to generate fine-grained details by attending to relevant words in the description. 
These models generate high fidelity images on single domain datasets (\eg, birds~\cite{WahCUB_200_2011} and flowers~\cite{Nilsback08}), but struggle on complex scenes with many objects---such as those in MS-COCO~\cite{LinMBHPRDZ14}. Recent methods~\cite{HongYCL18,objgan19,HinzHW19,trecs2020} propose object-driven, hierarchical approaches that explicitly model object instances within an image. Given the text description, they first infer a semantic layout (\eg, object bounding boxes, segmentation masks, or a combination), and then generate an image from the layout. 
These hierarchical methods are cumbersome to apply to real-world scenarios; generation becomes a multi-step process (box-to-mask-to-image), and the model requires much more fine-grained object labels to train.

We study contrastive learning in the context of text-to-image synthesis and demonstrate that a simple one-stage GAN \textit{without} object-level annotation can outperform prior object-driven and multi-stage approaches. Besides generating realistic images, we also hope (1) the image should holistically match the description; (2) generated images should match real images when they are conditioned on the same description; (3) individual image regions should be recognizable and consistent with words in the sentence. To fulfill these desiderata and achieve strong language alignment, we propose to maximize the mutual information between the corresponding pairs through contrastive learning.
Our method, the Cross(X)-Modal Contrastive Generative Adversarial Network (\CLGAN), uses image to sentence, image region to word, and image to image contrastive losses to enforce alignment between generated images and their captions (Fig.~\ref{fig:overview}). Our primary contributions include:

\begin{itemize}[noitemsep]
    \item We propose \CLGAN, a simple one-stage GAN that employs several contrastive losses. \CLGAN produces dramatic improvements over previous models, e.g. reducing FID~\cite{FID} from 24.70 to 9.33 on \coco and from 48.70 to 14.12 on \cocoln (the \coco portion of Localized Narratives~\cite{pont2019connecting}).
    \item We conduct thorough human evaluations comparing \CLGAN to three recent models. These show that people prefer \CLGAN 77.3\% of the time for image realism, and 74.1\% for image-text alignment.
    \item We establish a strong benchmark on the challenging \lnoi (Open Images subset of Localized Narratives). To the best of our knowledge, this is the first text-to-image results training and testing on the diverse images and descriptions for Open Images.
    \item We conduct a thorough analysis of contrastive losses used in \CLGAN to provide general modeling insights for contrastive learning in conditional GANs.
\end{itemize}

\noindent
\clgan consistently produces images that are more coherent and detailed than previous models. In addition to greater realism (with clearer, more delineated objects), they better capture the full image description, including the presence of named objects and background compositions.



\cutsectionup
\section{Related Work}
\label{sec:related_work}
\cutsectiondown

\paragraph{Text-to-image synthesis}
Generating images from text descriptions has been quickly improved with deep generative models, including pixelCNN~\cite{Oord16, Reed17parallel}, approximate Langevin sampling~\cite{NguyenYBDC17}, variational autoencoders (VAEs)~\cite{KingmaW14, Gregor15DRAW} and  Generative Adversarial Networks (GANs)~\cite{goodfellow2014generative, reed2016generative}. GAN-based models in particular have shown better sample quality~\cite{Han17,zhang2018photographic, xu18, zhu2019dm,yin2019semantics,li2019control,Tan2019SemanticsEnhancedAN, qiao2019learn, lao2019dual}. GAN-INT-CLS~\cite{reed2016generative} was the first to use conditional GANs for text to image generation.  StackGAN~\cite{Han17, Han17stackgan2} improves this with a coarse-to-fine framework that progressively generates images at different resolutions for high-resolution synthesis. AttnGAN~\cite{xu18} introduces cross-modal attention to better capture details. DM-GAN~\cite{zhu2019dm} adaptively refines generated images with a memory module that writes and reads text and image features. MirrorGAN~\cite{qiao2019mirrorgan} enforces text-image consistency via caption generation on the generated images. SD-GAN~\cite{yin2019semantics} proposes word-level conditional batch normalization and dual encoder structure with triplet loss to improve text-image alignment. Compared with the triplet loss, our contrastive loss does not require mining for informative negatives and thus lowers training complexity. CP-GAN~\cite{liang2019cpgan} proposes an object-aware image encoder and fine-grained discriminator. Its generated images obtain high Inception Score~\cite{salimans2016improved}; however, we show it performs poorly when evaluated with the stronger FID~\cite{FID} metric and in human evaluations (see Sec.~\ref{sec:compareprevious}). To create a final high resolution image, these approaches rely on multiple generators and discriminators to generate images at different resolutions. Others have proposed hierarchical models that explicitly generate different objects after inferring semantic layouts \cite{HongYCL18,HinzHW19,trecs2020}. A drawback of these is that they need fine-grained object labels (\eg, object bounding boxes or segmentation maps), so generation is a multi-step process. Compared to these multi-stage and multi-step frameworks, our proposed \CLGAN only has a single generator and discriminator trained end-to-end, and it generates much higher quality images.

\cutparagraphup
\paragraph{Contrastive learning and its use in GANs}
Contrastive learning is a powerful scheme for self-supervised representation learning~\cite{oord2018representation, he2019moco, chen2020simple, wu2018unsupervised}. It enforces consistency of image representations under different augmentations by contrasting positive pairs with negative ones. It has been explored under several adversarial training scenarios~\cite{lee2020infomax, zhao2020image, deng2020disentangled, qiao2018geometry}. Cntr-GAN~\cite{zhao2020image} uses a contrastive loss as regularization on image augmentations for unconditional image generation. 
ContraGAN~\cite{kang2020contrastive} explores contrastive learning for class-conditional image generation. DiscoFaceGAN~\cite{deng2020disentangled} adds contrastive learning to enforce disentanglement for face generation. CUT~\cite{park2020contrastive} proposes patch-based contrastive learning for image-to-image translation by using positive pairs from the same image location in input and output images. 
Unlike prior work, we use intra-modality (image-image) and inter-modality (image-sentence and region-word) contrastive learning in text-to-image synthesis (Fig.~\ref{fig:overview}).


\begin{figure*}
    \centering
    \includegraphics[width=\linewidth]{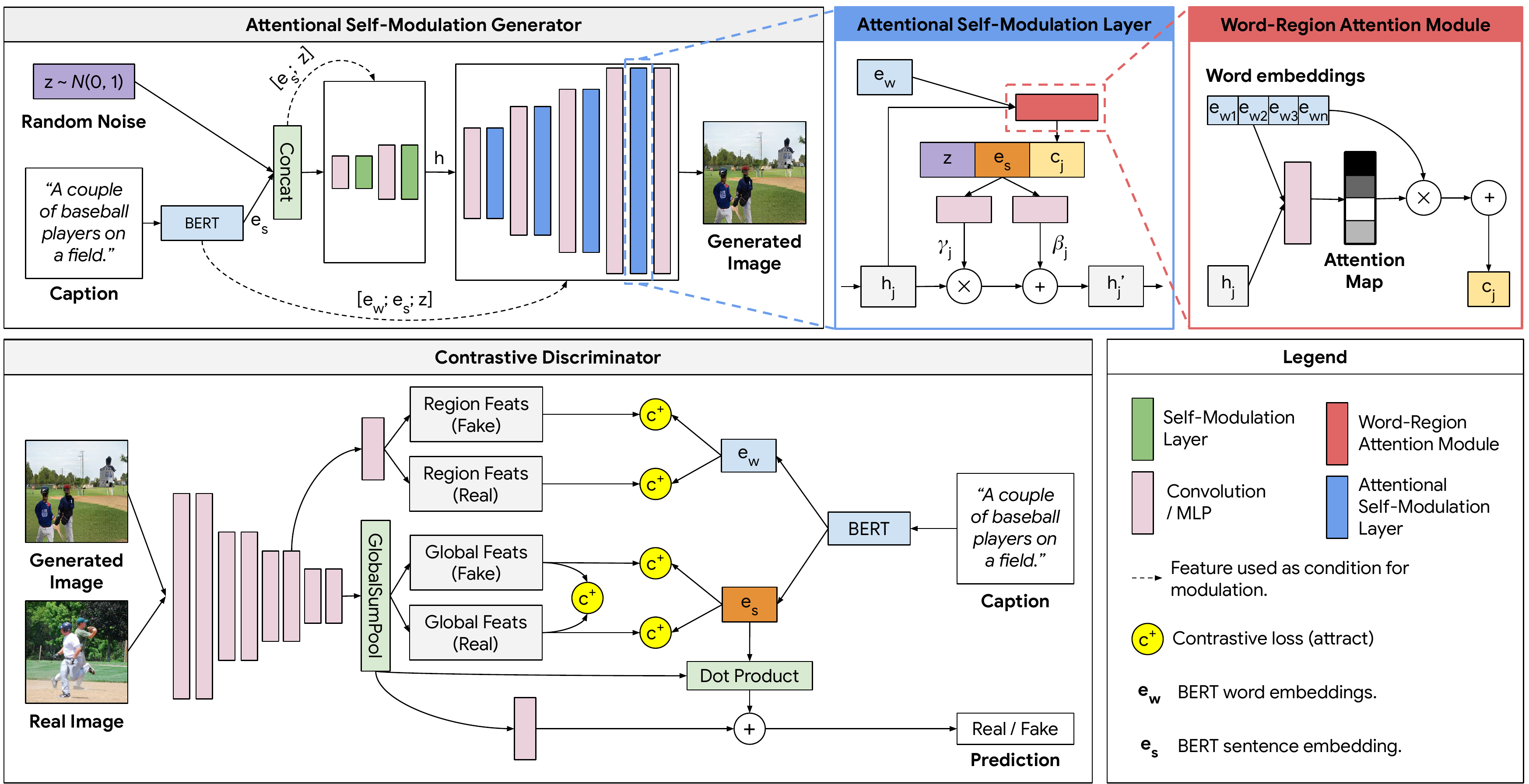}
    \vspace{-10pt}
    \caption{Overview of the proposed \CLGAN.}
    \label{fig:model}
\end{figure*}

\cutsectionup
\section{Preliminaries}

\cutsubsectionup
\subsection{Contrastive Representation Learning}
\label{sec:contrastive}
\cutsubsectiondown
\cutsubsectiondown
Contrastive learning aims to learn useful features given different views of data~\cite{Tian0PKSI20}. For example, note that $v_1$ and $v_2$ are two random variables to represent two different views of data. Feature representations are learned by measuring the mutual dependence $I(v_1; v_2)$ between these two variables. As directly maximizing the mutual information is challenging~\cite{paninski2003estimation, belghazi2018mutual,song2019understanding}, the InfoNCE loss~\cite{oord2018representation} was proposed to maximize a lower bound of the mutual information $I(v_1; v_2)$
Specifically, given a query sample $v_{1,i}$, minimizing the InfoNCE loss is to score the matching positive sample $v_{2,i} \sim p(v_2|v_{1,i})$ higher than $M{-}1$ negative samples $v_{2,j} \sim p(v_2)$. The overall objective can be summarized as follows:
\begin{equation} \label{equation:loss_nce}
\begin{split}
& I(v_1;v_2) \geq \log(M) - \mathcal{L}_{NCE},\\
\text{where } \mathcal{L}_{NCE} &= -\mathbb{E}\Bigg[ \log \frac{\exp(\mathcal{S}(v_{1,i}, v_{2,i}))}{\sum_{j=1}^{M}\exp(\mathcal{S}(v_{1,i}, v_{2,j}))}\Bigg]. \\
\end{split}
\end{equation}
Here, $\mathcal{S}(\cdot, \cdot)$ is the score function, which usually has two parameterized feature encoders for $v_1$ and $v_2$.  The encoders can share parameters if $v_1$ and $v_2$ are from the same modality. 
There are many ways to construct $v_1$ and $v_2$: different image augmentations \cite{he2019moco,chen2020simple}; spatially adjacent image patches \cite{oord2018representation}; a video as $v_1$ and its aligned audio as $v_2$ for video representation learning \cite{morgado2020audio, chung2019perfect}. 


\cutsubsectionup
\subsection{Generative Adversarial Networks (GANs)} 
\cutsubsectiondown

\label{sec:gans}
GANs \cite{goodfellow2014generative} are generative models that employ both a generator and a discriminator. The generator $G$ maps a latent variable $z{\sim}p{(z)}$ (usually sampled from a Gaussian distribution) to a real data distribution $p_\text{data}$. The discriminator $D$ is trained to distinguish whether inputs are synthesized by $G$ or sampled from real data. The generator $G$ is trained to synthesize images that the discriminator will
classify as real.


A large amount of work has focused on designing the adversarial objective to improve training~\cite{goodfellow2014generative, WGAN, Mao2016, Salimans18, lim2017, Tran2017}. A notable example is the hinge loss:

\begin{equation}
\begin{split}
\mathcal{L}_D=&-\mathbb{E}_{x \sim p_\text{data}}\left[\min(0, -1+D(x))\right] \\
   &- \mathbb{E}_{z \sim p(z)}\left[\min(0, -1-D(G(z)))\right], \\
\mathcal{L}_G=&- \mathbb{E}_{z \sim p(z)}\left[D(G(z))\right]. 
\end{split} \label{eq:hinge_gan}
\end{equation}

\noindent
The hinge loss has been used in state-of-the-art GANs for image generation~\cite{Miyato18a,SAGAN, BIGGAN, zhang2019consistency}. For conditional GANs, the generator and the discriminator are provided with an additional condition $c$, yielding $G(z, c)$ and $D(x, c)$. For conditional generation, the generated sample should be both realistic and also match the condition $c$.

\cutsectionup
\section{Method} \label{sec:method}
\cutsectiondown

We describe the losses and components of \clgan below. See Fig.~\ref{fig:model} for an overview.

\cutsubsectionup
\subsection{Contrastive Losses for Text-to-Image Synthesis}
\cutsubsectiondown



Text-to-image synthesis is a conditional generation task. Generated images should both be realistic and well-aligned with a given description. To achieve this, we propose to maximize the mutual information between the corresponding pairs: (1) image and sentence, (2) generated image and real image with the same description, and (3) image regions and words.
Directly maximizing mutual information is difficult (see Sec.~\ref{sec:contrastive}), so we maximize the lower bound of the mutual information by optimizing contrastive (\ie, InfoNCE) losses. 

\cutparagraphup
\paragraph{Image-text contrastive loss.} Given an image $x$ and its corresponding description $s$, we define the score function following previous work in contrastive learning~\cite{he2019moco,chen2020simple, oord2018representation}:
\begin{align} \label{eq:score_function}
\ssent(x, s) = \cos(f_{\text{img}}(x), f_{\text{sent}}(s))/\tau,
\end{align}
where $\cos(u, v) = u^T v / \Vert u \Vert \Vert v \Vert$ denotes cosine similarity, and $\tau$ denotes a temperature hyper-parameter. $f_{\text{img}}$ is an image encoder to extract the overall image feature vector and $f_{\text{sent}}$ is a sentence encoder to extract the global sentence feature vector. This maps the image and sentence representations into a joint embedding space $\mathbb{R}^D$. The contrastive loss between image $x_i$ and its paired sentence $s_i$ is computed as:
\begin{align}\label{eq: sentence_contrastive}
\lsent (x_i, s_i) = - \log \frac{\exp(\cos(f_{\text{img}}(x_i), f_{\text{sent}}(s_i))/\tau)}{\sum_{j=1}^{M}\exp(\cos(f_{\text{img}}(x_i), f_{\text{sent}}(s_j))/\tau)}. 
\end{align}
This form of contrastive loss is also known as the normalized temperature-scaled cross entropy loss (\textit{NT-Xent})~\cite{chen2020simple}.

\cutparagraphup
\paragraph{Contrastive loss between fake and real images with shared description.}
This contrastive loss is also defined with \textit{NT-Xent}. The main difference is that a shared image encoder $f_{\text{img}}^{\prime}$ extracts features for both real and fake images. The score function between two images is $\simg(x, \tilde{x})=\cos(f_{\text{img}}^{\prime}(x), f_{\text{img}}^{\prime}(\tilde{x}))/\tau$.
The image-image contrastive loss between real image $x_i$ and generated image $G(z_i, s_i)$ is:
\begin{equation}\label{eq: image_contrastive}
\begin{split}
\limg (x_i, G(z_i, s_i)) = - \log \frac{\exp(\simg(x_i, G(z_i, s_i)))}{\sum_{j=1}^{M}\exp(\simg(x_i, G(z_j, s_j)))}.
\end{split}
\end{equation}

\cutparagraphup
\paragraph{Contrastive loss between image regions and words.}
Individual image regions should be consistent with corresponding words in an input description. 
We use attention~\cite{xu18} to learn connections between regions in image $x$ and words in sentence $s$, without requiring fine-grained annotations that align words and regions. We first compute the pairwise cosine similarity matrix between all words in the sentence and all regions in the image; then, we compute the soft attention $\alpha_{i,j}$ for word $w_i$ to region $r_j$ as:
\begin{align}\label{equation: attn_region}
\alpha_{i,j} = \frac{\exp(\rho_1 \cos(f_{\text{word}}(w_i), f_{\text{region}}(r_j)))}{\sum_{h=1}^{R}\exp( \rho_1\cos(f_{\text{word}}(w_i),f_{\text{region}}(r_h)))},
\end{align}
where $f_{\text{word}}$ and $f_{\text{region}}$ represent word and region feature encoders respectively, $R$ is the total number of regions in the image and $\rho_1$ is a sharpening hyper-parameter to reduce the entropy of the soft attention. The aligned region feature for the $i^{th}$ word is defined as $c_i = \sum_{j=1}^{R} \alpha_{i,j} f_{\text{region}}(r_j)$. The score function between all the regions in image $x$ and all words in sentence $s$ can then be defined as:
\small
\begin{align}
\sword (x, s) = \log \Big(\sum_{h=1}^{T} \exp(\rho_2 \cos(f_{\text{word}}(w_h), c_h))\Big)^{\frac{1}{\rho_2}}/\tau,
\end{align}
\normalsize
where $T$ is the total number of words in the sentence. $\rho_2$ is a hyper-parameter that determines the weight of the most aligned word-region pair, \eg, as $\rho_2 \rightarrow \infty$, the score function approximates to $\max_{h=1}^{T} \cos(f_{\text{word}}(w_h), c_h)$. Finally the contrastive loss between the words and regions in image $x_i$ and its aligned sentence $s_i$ can be defined as:
\begin{align}
\lword (x_i, s_i) = - \log \frac{\exp(\sword(x_i, s_i))}{\sum_{j=1}^{M}\exp(\sword(x_i, s_j))}.
\end{align}

\cutsubsectionup
\subsection{Attentional Self-Modulation Generator} \label{sec:generator}
\cutsubsectiondown

We propose a one-stage generator to directly generate the image at the desired resolution. This is much simpler than previous multi-stage generators that create images at multiple, different resolutions.
We first sample noise $z$ from a standard Gaussian distribution. We obtain the global sentence embedding $e_s$ and the word embeddings $e_w$ from a pretrained BERT~\cite{devlin2018bert} module.   
$e_s$ and $z$ are concatenated to form the global condition, which is passed through several up-sampling blocks (see appendix for details) to generate a $16 \times 16$ feature map.
The global condition is also used as the condition to calculate scale parameter $\gamma$ and shift parameter $\beta$ in conditional batch normalization layers.
This formulation is also known as self-modulation~\cite{chen2018self}.

The self-modulation layer improves consistency of the hidden feature with the conditional inputs, but it lacks finer details for each sub-region. To generate fine-grained, recognizable regions, we propose the \textit{attentional self-modulation layer}. Specifically, besides random noise $z$ and global sentence embedding $e_s$, we modify the attention mechanism~\cite{xu18} to calculate the word-context vector as the additional modulation parameter for each sub-region.
For the $j^{th}$ region with feature $h_j$, the word-context vector $c_j$ is:
\small \begin{equation}
\begin{split}
c_j = \sum_{i=1}^{T} \tilde{\alpha}_{j, i} e_{w_i}, \text{where } \tilde{\alpha}_{j,i} = \frac{\exp(\rho_0 \cos(e_{w_i}, h_j))}{\sum_{k=1}^{T}\exp(\rho_0 \cos(e_{w_k},h_j))},
\end{split} \label{eq:hinge_gan}
\end{equation} \normalsize
where $T$ is the total number of words in the sentence and $\rho_0$ is a sharpening hyper-parameter. Then, the modulated feature $h_j^{\prime}$ for the $j^{th}$ region can be defined as: 
\small
\begin{equation}
\begin{split}
h_j^{\prime} = \gamma_j (\text{concat}(z, e_s, c_j)) \odot \frac{h_j -\mu}{\sigma} + \beta_j(\text{concat}(z, e_s, c_j)),
\end{split} \label{eq:hinge_gan}
\end{equation}
\normalsize
where $\mu$ and $\sigma$ are the estimated mean and standard deviation from aggregating both batch and spatial dimensions. $\gamma_j(\cdot)$ and $\beta_j(\cdot)$ represent any function approximators; in our work we simply use linear projection layers. Further details of the generator can be found in the appendix.


\begin{algorithm}[t]
    \caption{\CLGAN Training Algorithm.} 
    \label{alg:main}
\begin{algorithmic}[1]
    \renewcommand{\algorithmicrequire}{\textbf{Input:}}
    \renewcommand{\algorithmicensure}{\textbf{Output:}}
    \Require generator and discriminator parameters $\theta_G, \theta_D$, contrastive loss coefficients $\lambda_1$, $\lambda_2$, $\lambda_3$, Adam hyperparameters $\beta_1, \beta_2$, generator and discriminator learning rate $lr_G$, $lr_D$,
    batch size $M$, number of discriminator iterations per generator iteration $N_D$
    
    \For{number of training iterations} 
    \For{$t=1,...,N_D$}            
    \State  Sample  $\{z_{i}\}_{i=1}^{M} \sim p(z)$ \State Sample $\{(x_i, s_i)\}_{i=1}^{M} \sim p_\text{data}(x,s)$ 
    \State $\el_{\text{sent}}^{\text{r}} \gets \frac{1}{M} \sum_{i=1}^M \el_{\text{sent}}(x_i, s_i)$
    \State $\el_{\text{word}}^{\text{r}} \gets \frac{1}{M} \sum_{i=1}^M \el_{\text{word}}(x_i, s_i)$
    \State $\el_{\text{GAN}}^{D} \gets -\frac{1}{M} \sum_{i=1}^M \min(0, -1+D(x_i, s_i)) -$
    \myindent{1.8} $\frac{1}{M} \sum_{i=1}^M \min(0, -1-D(G(z_i, s_i), s_i))$
    \State $\el_{D} \gets \el_{\text{GAN}}^{D} + \lambda_1 \el_{\text{sent}}^{\text{r}} + \lambda_2 \el_{\text{word}}^{\text{r}}$
    \State $\theta_D \gets \text{Adam}(\el_D, lr_{D}, \beta_1, \beta_2)$
    \EndFor
    \State  Sample  $\{z_{i}\}_{i=1}^{M} \sim p(z)$, $\{(x_i, s_i)\}_{i=1}^{M} \sim p_\text{data}(x,s)$ 
    \State $\el_{\text{sent}}^{\text{f}} \gets \frac{1}{M} \sum_{i=1}^M \el_{\text{sent}}(G(z_i, s_i), s_i)$
    \State $\el_{\text{word}}^{\text{f}} \gets \frac{1}{M} \sum_{i=1}^M \el_{\text{word}}(G(z_i, s_i), s_i)$
    \State $\el_{\text{img}} \gets \frac{1}{M} \sum_{i=1}^M \el_{\text{img}}(G(z_i, s_i), x_i)$
    \State $\el_{\text{GAN}}^{G} \gets \frac{1}{M} \sum_{i=1}^M -(D(G(z_i, s_i), s_i))$
    \State $\el_{G} \gets \el_{\text{GAN}}^{G} + \lambda_1 \el_{\text{sent}}^{\text{f}} + \lambda_2 \el_{\text{word}}^{\text{f}} + \lambda_3 \el_{\text{img}}$
    \State $\theta_{G} \gets  \text{Adam}(\el_G, lr_G, \beta_1, \beta_2)$
    \EndFor
\end{algorithmic}
\end{algorithm}

\cutsubsectionup
\subsection{Contrastive Discriminator} \label{sec:discriminator}
\cutsubsectiondown

Our proposed discriminator has two roles: (1) to act as a critic to determine whether an input image is real or fake, and (2) to act as an encoder to compute global image and region features for the contrastive loss. The image is passed through several down-sampling blocks until its spatial dimensions are reduced to $16{\times}16$ (see Fig.~\ref{fig:model}, bottom left). Then, a $1{\times}1$ convolution is applied to obtain region features, where the feature dimensions are consistent with the dimensions of the word embedding. The original image feature is fed through two more down-sampling blocks and a global pooling layer. Finally, a projection head computes the logit for the adversarial loss, and a separate projection head computes image features for the image-sentence and image-image contrastive loss.
Note that it is important to only use the \textit{real images} and their descriptions to train these discriminator projection heads. The reason is that the generated images are sometimes not recognizable, especially at the start of training. Using such generated image and sentence pairs hurts the training of the image feature encoder projection heads. Therefore, the contrastive losses from \textit{fake images} are only applied to the generator. In addition to the discriminator projection layers, we use a pretrained VGG network~\cite{VGGNet} as an image encoder for an additional supervisory image-image contrastive loss (see Sec.~\ref{sec:componentanalysis}). Algorithm~\ref{alg:main} summarizes the \CLGAN training procedure. For simplicity, we set all contrastive loss coefficients ($\lambda_1, \lambda_2, \lambda_3$ in Algorithm~\ref{alg:main}) to 1.0 in our experiments.

\cutsectionup
\section{Evaluation}

\cutsubsectionup
\subsection{Data} \label{sec:datasets}
\cutsubsectiondown

We perform a comprehensive evaluation of \CLGAN on three challenging datasets (summarized in Table~\ref{tab:dataset}).

\textbf{\coco}~\cite{LinMBHPRDZ14} is commonly used for text-to-image synthesis. Each image is paired with 5 short captions. We follow most prior work to use the 2014 split (\cocoold) for evaluation.


\begin{table}[bt]
\begin{center}
\small
\resizebox{\linewidth}{!}{%
\begin{tabular}{lcccccccc}
\hline
 \multirow{2}{4em}{\textbf{Dataset}}&\multicolumn{2}{c}{\textbf{\cocoold}}  & \multicolumn{2}{c}{\textbf{\cocoln}} & \multicolumn{2}{c}{\textbf{\lnoi}}  \\
\cline{2-7}
 &train &val &train &val &train &val\\
\hline
 \#samples & 82k & 40k & 134k & 8k & 507k & 41k\\
\hline
caption/image& \multicolumn{2}{c}{5} & \multicolumn{2}{c}{1} & \multicolumn{2}{c}{1} \\
\hline
avg. caption length&  \multicolumn{2}{c}{10.5} & \multicolumn{2}{c}{42.1} & \multicolumn{2}{c}{35.6} \\
\hline
\end{tabular}
}
\end{center}
\vspace{-10pt}
\caption{Statistics of datasets.}
\vspace{-10pt}
\label{tab:dataset} 
\end{table}


Localized Narratives~\cite{pont2019connecting} contains long form image descriptions for several image collections. We benchmark results on \textbf{\cocoln}, which contains \textit{narratives} for images in the 2017 split of \coco (\coconew). Narratives are four times longer than \coco captions on average and they are much more descriptive (see Figure~\ref{fig:compare_coco}).
Narratives also contain disfluencies since they are spoken and then transcribed. These factors make text-to-image synthesis for \cocoln much more challenging than \coco.

We also train and evaluate using \textbf{\lnoi}, the \oi~\cite{kuznetsova2018open} split of Localized Narratives. Its images are both diverse and complex (8.4 objects on average). \lnoi is also much larger than \coco and \cocoln (see Table~\ref{tab:dataset}). To the best of our knowledge, we are the first to train and evaluate a text-to-image generation model for \oi. \CLGAN is able to generate high quality results, and sets a strong benchmark for this very challenging task.


\begin{table*}
\begin{center}
\resizebox{0.7\linewidth}{!}{%
\begin{tabular}{l@{\hspace{8mm}}c@{\hspace{4mm}}c@{\hspace{4mm}}c@{\hspace{4mm}}c@{\hspace{4mm}}c@{\hspace{4mm}}}
\hline
\textbf{Model}  & \textbf{IS} $\uparrow$ & \textbf{FID} $\downarrow$ & \textbf{R-prec (CC) $\uparrow$} & \textbf{SOA-C} $\uparrow$ & \textbf{SOA-I} $\uparrow$ \\
\hline
Real Images & 34.88 & 6.09 & 69.36 & 74.97 & 80.84 \\
\hline
AttnGAN \cite{xu18} & 23.61 & 33.10 & - & 25.88 & 39.01 \\
Obj-GAN \cite{objgan19} & 24.09 & 36.52 & - & 27.14 & 41.24 \\
DM-GAN \cite{zhu2019dm} & 32.32 & 27.34 & - & 33.44 & 48.03 \\ 
OP-GAN \cite{hinz2020semantic} & 27.88 & 24.70 & 49.80 & 35.85 & 50.47 \\
SD-GAN \cite{yin2019semantics} & 35.69 & \pz29.35$^\dagger$ & 51.68 & - & - \\
CP-GAN \cite{liang2019cpgan} & \textbf{52.73} & \pz55.82$^\ddagger$ & 59.05 & \textbf{77.02} & \textbf{84.55} \\
\CLGAN (ours) & 30.45 & \pz\textbf{9.33} & \textbf{71.00} & 50.94 & 71.33 \\
\hline
\end{tabular}
}
\end{center}
\vspace{-10pt}
\caption{Comparison of \CLGAN with previous models on \cocoold. \textit{R-prec (CC)} are R-precision scores computed from a model trained on Conceptual Captions (see Sec.~\ref{sec:evaluation}). $^\dagger$ indicates scores computed from images shared by the original paper authors, and $^\ddagger$ indicates scores computed from images generated from the open-sourced models.}
\label{tab:compare_others}
\vspace*{-0.2in}
\end{table*}


\cutsubsectionup
\subsection{Evaluation Metrics} \label{sec:evaluation}
\cutsubsectiondown

Following previous work, we report validation results by generating images for 30,000 random captions\footnote{We oversample the images and captions if there are less than 30,000 samples in the validation set.}. We evaluate comprehensively using several measures.

\cutparagraphup
\paragraph{Image quality.} We use standard automated metrics for assessing image quality. \textit{Inception Score (\IS)}~\cite{salimans2016improved} calculates \textit{KL}-divergence between the conditional class distribution and the marginal class distribution given a pre-trained image classifier.  \textit{Fr\'echet Inception Distance (\FID)}~\cite{FID} is the Fr\'echet distance between two multivariate Gaussians fit to Inception~\cite{Szegedy2016} features of generated and real images. While \IS and \FID have both been shown to correlate with human judgements of generated image quality, IS is likely less informative as it overfits easily and can be manipulated to achieve much higher scores using simple tricks \cite{ISISSUES,hinz2020semantic}. This is further emphasized by our results (Sec. \ref{sec:compareprevious}) showing that FID correlates better with human judgments of realism. 



\cutparagraphup
\paragraph{Text-Image Alignment.} Following previous work~\cite{xu18,objgan19}, we use \textit{R-precision} to assess whether a generated image can be used to retrieve its conditioning description. However, we notice that previous work computes R-precision using image-text encoders from AttnGAN~\cite{xu18}, and many others use these encoders as part of their optimization function during training.
This skews results: many generated models report R-precision scores significantly higher than real images. To alleviate this, we use an image-text dual-encoder\footnote{This model will be publicly released to facilitate future evaluations.}~\cite{crisscross_2020} pretrained on \textit{real images} in the Conceptual Captions dataset~\cite{sharma2018conceptual}, which is disjoint from \coco. We find that computing R-precision with independent encoders better correlates with human judgments.

Caption retrieval metrics assess whether the entire image matches the caption. In contrast, \textit{Semantic Object Accuracy (SOA})~\cite{hinz2020semantic} evaluates the quality of individual regions and objects within an image. 
Like previous work, we report SOA-C (\ie, the percentage of images per class in which a desired object is detected) and SOA-I (\ie, the percentage of images in which a desired object is detected). Further details of SOA can be found in~\cite{hinz2020semantic}. SOA was originally designed for \cocoold, and can take very long to compute as it requires generating multiple samples for each \coco class label. We use the official code to compute the metrics reported in Table~\ref{tab:compare_others}, but approximate results for \cocoln and other ablation experiments where we compute results over 30,000 random samples.


\cutparagraphup
\paragraph{Human evaluation.} Automated metrics are useful while iterating on models during experimentation, but they are no substitute for human eyes. We conduct thorough human evaluations on generated images from 1000 randomly selected captions. For each caption, we request 5 independent human annotators to rank the generated images from best to worst based on (1) realism, and (2) language alignment. 

\begin{figure}
    \centering
    \includegraphics[width=\linewidth]{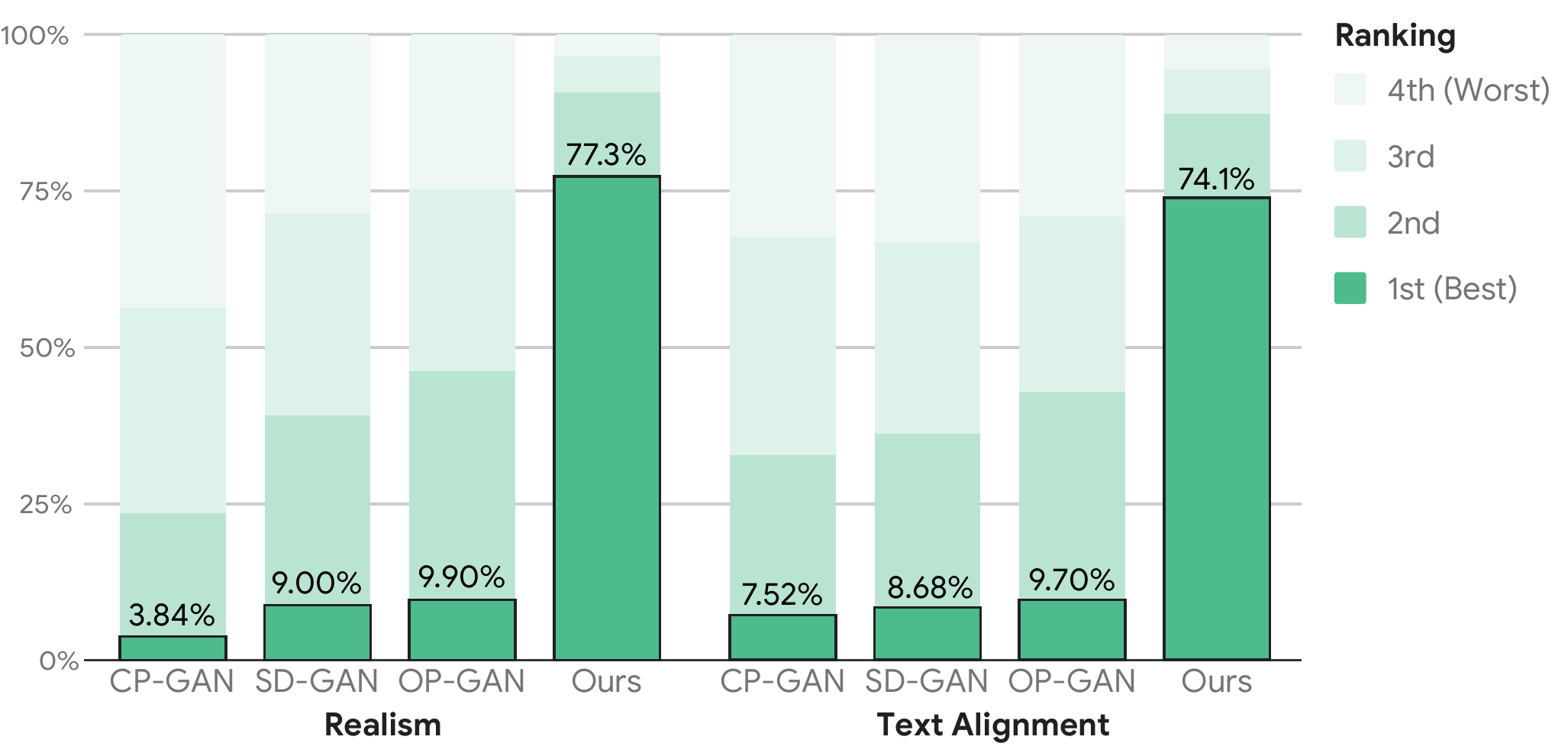}
    \caption{Human evaluation on \cocoold for image quality and text alignment. Annotators rank (anonymized and order-randomized) generated images from best to worst.}
    \label{fig:humanevals}
\end{figure}

\cutsectionup
\section{Experiments} \label{sec:experiment}


\cutsubsectionup
\subsection{Results} \label{sec:compareprevious}
\cutsubsectiondown

\begin{figure*}
\small
\begin{center}
\begin{tabular}{>{\arraybackslash}m{0.1\linewidth}@{\hskip 0.5em}c@{\hskip 0.1em}c@{\hskip 0.1em}c@{\hskip 0.1em}c@{\hskip 0.2em}|>{\arraybackslash}m{0.2\linewidth}@{\hskip 0.5em}c@{\hskip 0.1em}c@{\hskip 0.1em}c@{\hskip 0.1em}}
\textbf{\coco Caption} & \textbf{OP-GAN} & \textbf{SD-GAN} & \textbf{CP-GAN} & \textbf{\CLGAN} & \textbf{\cocoln Caption} & \textbf{AttnGAN} & \textbf{TReCS} & \textbf{\CLGAN} \\ \midrule \addlinespace[-0.05em]
\compareexamplesmall{a green train is coming down the tracks}{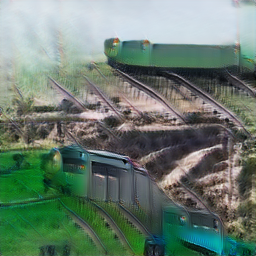} &
\compareexamplelnnooriginal{There is a group of people. They are standing on ski board. They are smiling. They are holding a sticks. In the center of the person is wearing a helmet. On the right side ...}{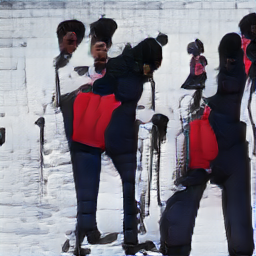}{000000470779} \\
\compareexamplesmall{A group of skiers are preparing to ski down a mountain.}{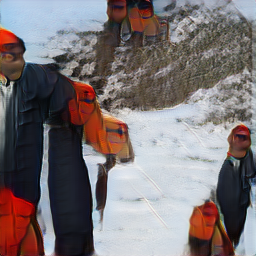} &
\compareexamplelnnooriginal{In this image I can see people are sitting on chairs. I can also see few of them are wearing shades. Here I can see few more chairs and tables. On this table I can see food ...}{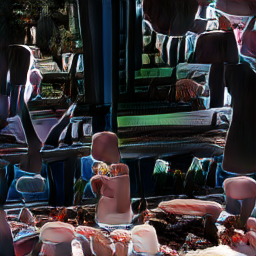}{000000470924} \\
\compareexamplesmall{A small kitchen with low a ceiling}{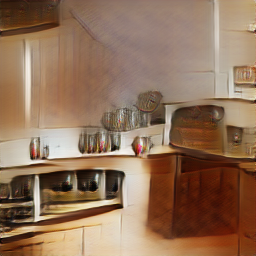} &
\compareexamplelnnooriginal{This picture shows an inner view of a restroom we see a wash basin with tap and a mirror on the wall and we see a light on it and we see a toilet seat and a frame on the wall and ...}{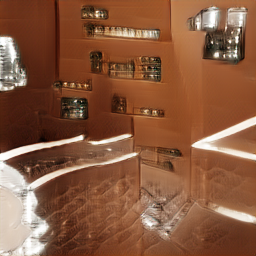}{000000489091} \\\addlinespace[0.1em]
\compareexamplesmall{A child eating a birthday cake near some balloons.}{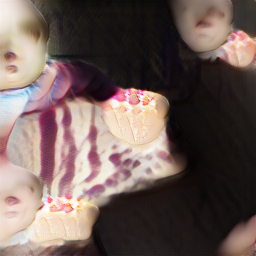} &
\compareexamplelnnooriginal{In this image we can see a red color train on the railway track. Here we can see platform}{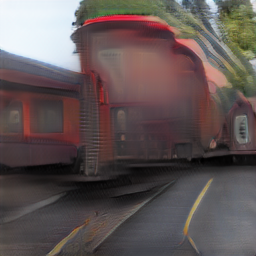}{000000539445} \\
\compareexamplesmall{A living area with a television and a table}{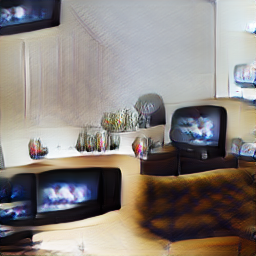} &
\compareexamplelnnooriginal{In this picture there are two members lying on the beach in the sand under an umbrella. There are some people standing here. In the background there is water}{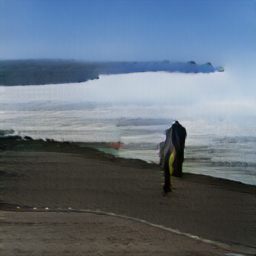}{000000262048} \\

\end{tabular}
\end{center}
\vspace{-10pt}
\caption{Generated images for selected examples from \cocoold and \cocoln. \CLGAN generated images are generally of much higher quality and depict clearer scenes. More random samples are available in the appendix.} \label{fig:compare_coco}
\end{figure*}

\paragraph{\cocoold.} Figure \ref{fig:humanevals} shows \textit{human evaluations} comparing \clgan to three recent strong models: CP-GAN~\cite{liang2019cpgan}, SD-GAN~\cite{yin2019semantics}, and OP-GAN~\cite{hinz2020semantic}. Given images (anonymized and randomly ordered) generated from the same caption by the four models, annotators are asked to rank them from best to worst. Realism and text alignment judgments are collected independently. \clgan is the clear winner on both: its output is ranked best in 77.3\% of realism comparisons, and 74.1\% of text alignment ones. OP-GAN is a distant second, at 9.90\% and 9.70\%, respectively. \clgan achieves this while being a simpler, one-stage model, whereas OP-GAN is multi-stage and needs object bounding boxes. Visual inspection of selected images (Fig. \ref{fig:compare_coco}) convincingly shows the large quality improvement. \CLGAN's images are much higher fidelity compared to others, and depict clearer objects and more coherent scenes. This also holds for more random samples (see appendix).

\begin{table}
\begin{center}
\resizebox{\linewidth}{!}{%
\begin{tabular}{lccccc}
\hline
\textbf{Model}  & \textbf{IS} $\uparrow$ & \textbf{FID} $\downarrow$  & \textbf{\rprec} $\uparrow$ & \textbf{SOA-C} $\uparrow$ & \textbf{SOA-I} $\uparrow$ \\
\hline
Real Images & 34.40 & 8.01 & 61.52 & 66.08 & 67.39  \\ 
\hline
AttnGAN~\cite{xu18} & 20.80 & 51.80 & 43.88 & - & -  \\ 
\textsc{TReCS}~\cite{trecs2020} & 21.30 & 48.70 & 37.88 & - & - \\ 
\CLGAN (ours) & \textbf{28.37} & \textbf{14.12} & \textbf{66.92} & 36.76 & 48.14  \\
\hline
\end{tabular}
}
\end{center}
\vspace{-10pt}
\caption{Comparison of \CLGAN on \cocoln. SOA metrics together with others are computed from 30,000 random examples.}
\label{tab:compare_ln} 
\vspace*{-0.2in}
\end{table}



Table~\ref{tab:compare_others} provides comprehensive \cocoold results for \textit{automated} metrics. \CLGAN dramatically 
improves FID from 24.70 to 9.33, a 62.2\% relative improvement over the next best model, OP-GAN~\cite{hinz2020semantic}. \CLGAN also outperforms others (71\% vs.~59\%) for R-precision computed with our \textit{independently trained} encoders, indicating a large improvement in fidelity of generated images to the captions they are conditioned on---and consistent with human judgments. Although CP-GAN achieves higher IS and SOA scores, both our human evaluations and visual inspection of randomly selected images indicates \CLGAN's image quality is much higher than CP-GAN's. This may be due to the issue that \IS and SOA do not penalize intra-class mode dropping (low diversity within a class)---a model that generates one ``perfect" sample for each class can achieve good scores on \IS and SOA. Our findings are consistent with other works~\cite{objgan19,ISISSUES}, which suggest that \FID may be a more reliable metric for measuring text-to-image synthesis quality.


\cutparagraphup
\paragraph{\cocoln.} Localized Narratives~\cite{pont2019connecting} contains much longer descriptions, which increases the difficulty of text-to-image synthesis (see Sec. \ref{sec:datasets}). Table~\ref{tab:compare_ln} shows that \CLGAN provides massive improvements over prior work. Compared to TReCS~\cite{trecs2020}, \clgan improves IS and FID, by 7.07 and 34.58 (absolute), respectively. It also improves R-precision by 23.04\% absolute over AttnGAN~\cite{xu18}, indicating much better text alignment. This is supported by qualitative comparison of randomly selected outputs:
\CLGAN's images are decisively clearer and more coherent (see Fig.~\ref{fig:compare_coco}).
We stress that TReCS exploits \cocoln's mouse trace annotations---incorporating this training signal in \clgan in future should further boost performance.


\begin{table}[t]
\begin{center}
\resizebox{\linewidth}{!}{%
\begin{tabular}{cccccccc}
\hline
\textbf{S} & \textbf{W} & \textbf{I} & \textbf{IS} $\uparrow$ & \textbf{FID} $\downarrow$  & \textbf{R-prec} $\uparrow$ & \textbf{SOA-C} $\uparrow$ & \textbf{SOA-I} $\uparrow$ \\
\hline
 \multicolumn{3}{c}{Real Images~\cite{hinz2020semantic}} & 34.88 & \pz 6.09 & 69.36 & 76.17 & 80.12 \\
\hline
 & & & 15.89 & 39.28 & 21.41 & 8.99 & 25.72  \\
 \checkmark & & & 23.50 & 19.25 & 53.57 & 24.57 & 45.41  \\ 
 & \checkmark & & 20.72 & 24.38 & 44.42 & 20.50 & 39.12  \\ 
 & & D &  18.90 & 29.71 & 31.16 & 12.73 & 30.89 \\ 
 & & VGG & 21.54 & 39.58 & 35.89 & 17.41 & 35.08  \\ 
 & & D + VGG & 23.61 & 21.14 & 47.04 & 23.87 & 44.41  \\ 
 \checkmark & \checkmark & &  26.02 & 14.25 & 64.94 & 30.49 & 51.60  \\  
 \checkmark & \checkmark & D & 28.06 & 12.96 & 65.36 & 34.21 & 54.23  \\ 
 \checkmark & \checkmark & VGG & 30.55 & \textbf{11.12} & \textbf{70.98} & 39.36 & 59.10  \\ 
\checkmark & \checkmark & D + VGG & \textbf{30.66} & 11.93 & 69.86 & \textbf{39.85} & \textbf{59.78}  \\ 
\hline
\end{tabular}
}
\end{center}
\vspace{-10pt}
\caption{Ablation results with different contrastive losses on \cocoold. \textbf{S} indicates the sentence-image loss. \textbf{W} indicates the region-word loss. \textbf{I} indicates the image-image loss, where D represents using the discriminator to extract image features, and VGG represents using a pre-trained VGG network to extract image features.}
\vspace{-10pt}
\label{tab:ablation_2014}
\end{table}

\cutparagraphup
\paragraph{\lnoi.}
We train \clgan on \oi dataset, which is much more challenging than \coco due to greater diversity in images and descriptions. \clgan achieves an \IS of 24.90, FID of 26.91, and R-precision of 57.55, and manages to generate high quality images (see appendix). To the best of our knowledge, \clgan is the first text-to-image model trained and evaluated on \oi. Its strong automated scores establish strong benchmark results on this challenging dataset. 

\cutsubsectionup
\subsection{Ablations} \label{sec:componentanalysis}
\cutsubsectiondown

\begin{figure}
    \centering
    \includegraphics[width=\linewidth]{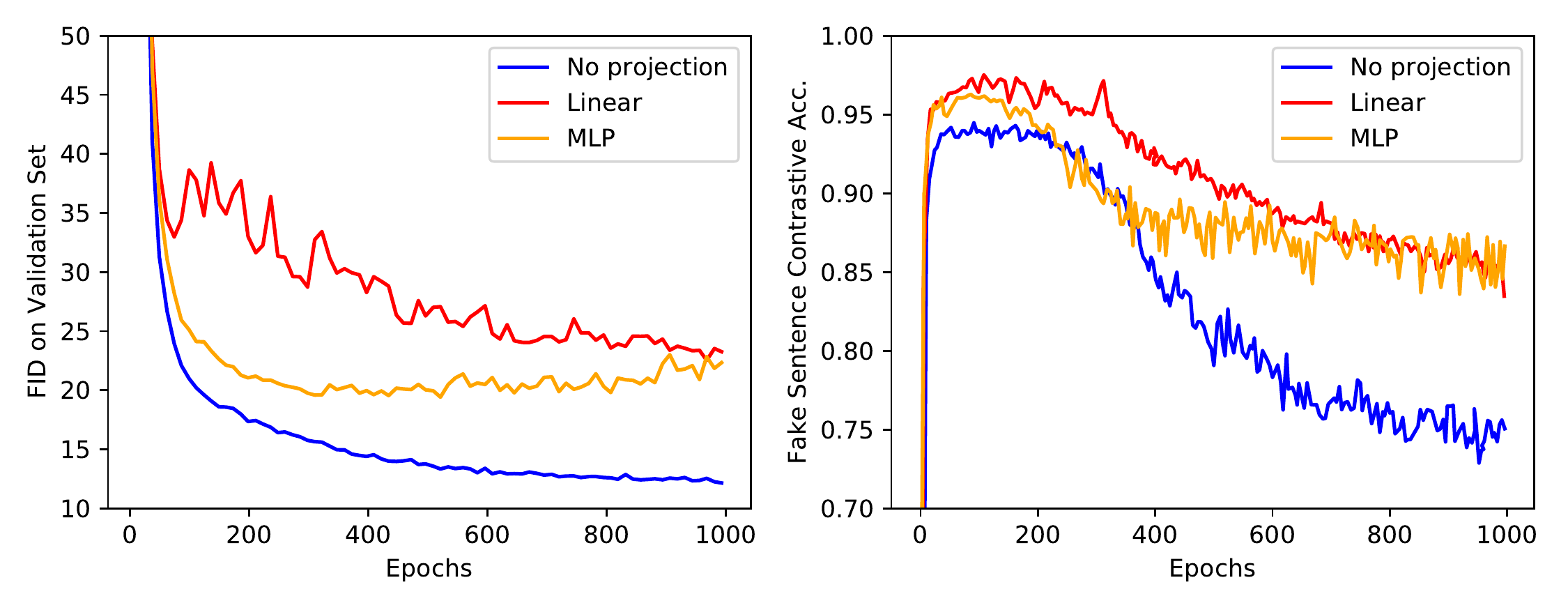}
\vspace{-15pt}
    \caption{Comparison between different contrastive heads.}
    \label{fig:contrasive_head_ablation}
\vspace{-15pt}
\end{figure}

We thoroughly evaluate the different components of \CLGAN and analyze their impact. Table~\ref{tab:ablation_2014} summarizes our ablations\footnote{All ablation results (Fig.~\ref{fig:contrasive_head_ablation}, Tables~\ref{tab:ablation_2014}, \ref{tab:ablation_modulation}, and \ref{tab:ablation_vgg}) are reported using metrics re-implemented in TensorFlow. SOA is approximated using 30,000 random samples. Ablation models use a reduced base channels dimension of 64. Implementation details are provided in the appendix.} on the \cocoold validation set. 
To study the effects of each contrastive loss component used in \CLGAN, we experiment with four 
losses: (1) image-sentence, (2) region-word, (3) image-image using discriminator features, and (4) image-image using VGG features. For (3), we use the discriminator encoder projection (indicated by D in Table~\ref{tab:ablation_2014}) to extract image features. For (4), we extract image features from a VGG-19 network~\cite{VGGNet} pretrained on ImageNet.

\cutparagraphup
\paragraph{Individual contrastive losses.}
Table~\ref{tab:ablation_2014} shows that using any of the contrastive losses improves all metrics compared to the baseline. During experimentation, we also found that including any contrastive loss greatly improves training stability. The largest improvements come from the \textit{inter-modal} image-sentence and region-word contrastive losses, which improve FID from 39.28 to 19.25 and 24.38, respectively. This is much larger compared to the image-image \textit{intra-modal} contrastive losses, \eg, including the loss from the discriminator feature encoder (D) only improves FID to 29.71. These ablations highlight the effectiveness of inter-modal contrastive losses: sentence and word contrastive losses each greatly improve the text-alignment metrics, as well as improving image quality.

\cutparagraphup
\paragraph{Combined contrastive losses.} Combining contrastive losses provides further gains. For example, using both image-sentence and region-word losses achieves better performance (FID 14.25) than alone (FID 19.25 and 24.38, respectively). This demonstrates that local and global conditions are complementary. Moreover, using both inter-modal losses (sentence and words) outperforms the intra-modal losses (D + VGG): FID scores are 14.25 and 21.14, respectively. These results further emphasize the effectiveness of cross-modal contrastive learning. Nevertheless, the \textit{inter-modal} and \textit{intra-modal} contrastive losses also complement each other:  the best \FID score comes from combining image-sentence, region-word, and image-image (VGG) losses. Performance on IS and text alignment further improves when using the image-image (D + VGG) loss. To obtain our final results (Table~\ref{tab:compare_others}), we train a model (with base channels dimension 96) using all 4 contrastive losses. 



\begin{table}
\begin{center}
\resizebox{\linewidth}{!}{%
\begin{tabular}{lccccc}
\hline
\textbf{Modulation}  & \textbf{IS} $\uparrow$ & \textbf{FID} $\downarrow$  & \textbf{\rprec} $\uparrow$ & \textbf{SOA-C} $\uparrow$ & \textbf{SOA-I} $\uparrow$ \\
\hline
Self-modulation & 28.98 & 13.59 & 64.65 & 35.18 & 55.54  \\
Attentional self-modulation & \textbf{30.66} & \textbf{11.93} & \textbf{69.86} & \textbf{39.85} & \textbf{59.78} \\
\hline
\end{tabular}
}
\end{center}
\vspace{-10pt}
\caption{Comparison of different modulation layers.}
\label{tab:ablation_modulation}
\end{table}

\begin{table}
\begin{center}
\resizebox{\linewidth}{!}{%
\begin{tabular}{lccccc}
\hline
\textbf{VGG Loss}  & \textbf{IS} $\uparrow$ & \textbf{FID} $\downarrow$  & \textbf{\rprec} $\uparrow$ & \textbf{SOA-C} $\uparrow$ & \textbf{SOA-I} $\uparrow$ \\
\hline
$l_2$ loss & 12.46 & 52.86 & 22.62 & 8.27 & 25.48  \\
Contrastive (InfoNCE) loss & \textbf{21.54} & \textbf{39.58} & \textbf{35.89} & \textbf{17.41} & \textbf{35.08} \\
\hline
\end{tabular}
}
\end{center}
\vspace{-10pt}
\caption{Comparison of different VGG losses.}
\vspace{-10pt}
\label{tab:ablation_vgg}
\end{table}

\paragraph{Deeper contrastive heads.} In unsupervised representation learning~\cite{chen2020simple,chen2020improved}, adding non-linear layers generally improves performance. To study this, we increase the depth of the projection head in the discriminator. 
Training curves for FID and contrastive accuracy~\cite{chen2020simple} on fake images are in Fig.~\ref{fig:contrasive_head_ablation}, across 1000 epochs. We find that using no additional projection layers gives the best \FID (12.61, compared to 19.42 of the 2-layer MLP). 
Moreover, we also find that the contrastive accuracy increases on fake images (from 76.56\% to 88.55\%) when more layers are added to the projection head. We posit that the discriminator overfits to the contrastive learning task in this configuration, resulting in poorer performance on the adversarial task as a critic and hence worse as a supervisory signal for the generator.


\cutparagraphup
\paragraph{Attentional Self-Modulation.} We compare two generator setups: (1) self-modulation layers~\cite{chen2018self} in all residual blocks, and (2) attentional self-modulation layers (see Sec.~\ref{sec:generator}) for blocks with input resolution larger than $16{\times}16$.  Table~\ref{tab:ablation_modulation} shows that the proposed attentional self-modulation layer outperforms self-modulation on all metrics.

\cutparagraphup
\paragraph{Loss types.} A frequently used loss function in generative models is the $l_2$ loss over VGG~\cite{VGGNet} outputs between fake images and corresponding real images. This is also commonly known as the perceptual loss~\cite{JohnsonAF16}. Table~\ref{tab:ablation_vgg} shows that contrastive losses outperform such perceptual losses. This demonstrates that repelling mismatched samples is more effective than simply pulling together aligned samples. Given this superior performance, replacing perceptual losses with contrastive losses may help other generative tasks.

\cutsectionup
\section{Conclusion} \label{sec:conclusion}
\cutsectiondown

In this work, we present a cross-modal contrastive learning framework to train GAN models for text-to-image synthesis. We investigate several cross-modal contrastive losses that enforce correspondence between image and text. With both human and automated evaluations on multiple datasets, \clgan establishes a marked improvement over previous models: it generates higher quality images that better match their input descriptions, including for long, detailed narratives. It does so while being a simpler, end-to-end model. We believe that these advances are strong leaps towards creative applications for image generation from natural language descriptions.

%

{\small
\bibliographystyle{ieee_fullname}
\bibliography{egbib}
}

\clearpage
\newpage

\pagebreak
\appendix
\noindent In this appendix, we share implementation details (Sec.~\ref{appendix:implementation}), architecture details (Sec.~\ref{appendix:architecture}), details about our human evaluation procedure (Sec.~\ref{appendix:human_evals}), and further qualitative results (Sec.~\ref{appendix:qual}).

\section{Implementation Details}
\label{appendix:implementation}
All models are implemented in TensorFlow 2.0. Spectral normalization is used for all convolutional and fully-connected layers in the discriminator. For training all models, we use the Adam optimizer with parameters $\beta_1=0.5$ and $\beta_2=0.999$. The learning rates for the generator and discriminator are set to $1e^{-4}$ and $4e^{-4}$ respectively. We use two discriminator training steps for each generator training step.
During validation, we report results from the generator with exponential moving averaged weights, with a decay rate of 0.999.

Models are trained with a batch size of 256. For reporting results in our paper, models are trained for 1000 epochs, and we report the scores corresponding to the checkpoint with the best FID score on the validation set. For reporting our main results, we train a model with base channel dimensions $ch = 96$ (see Table~\ref{tab:256_architecture}). For ablation experiments in the main paper, we train models with base channel dimensions $ch = 64$.

\section{Architecture Details}
\label{appendix:architecture}

\begin{figure*}[h]
    \centering
    \begin{subfigure}[t]{0.28\textwidth}
        \centering
        \includegraphics[width=1.0\textwidth]{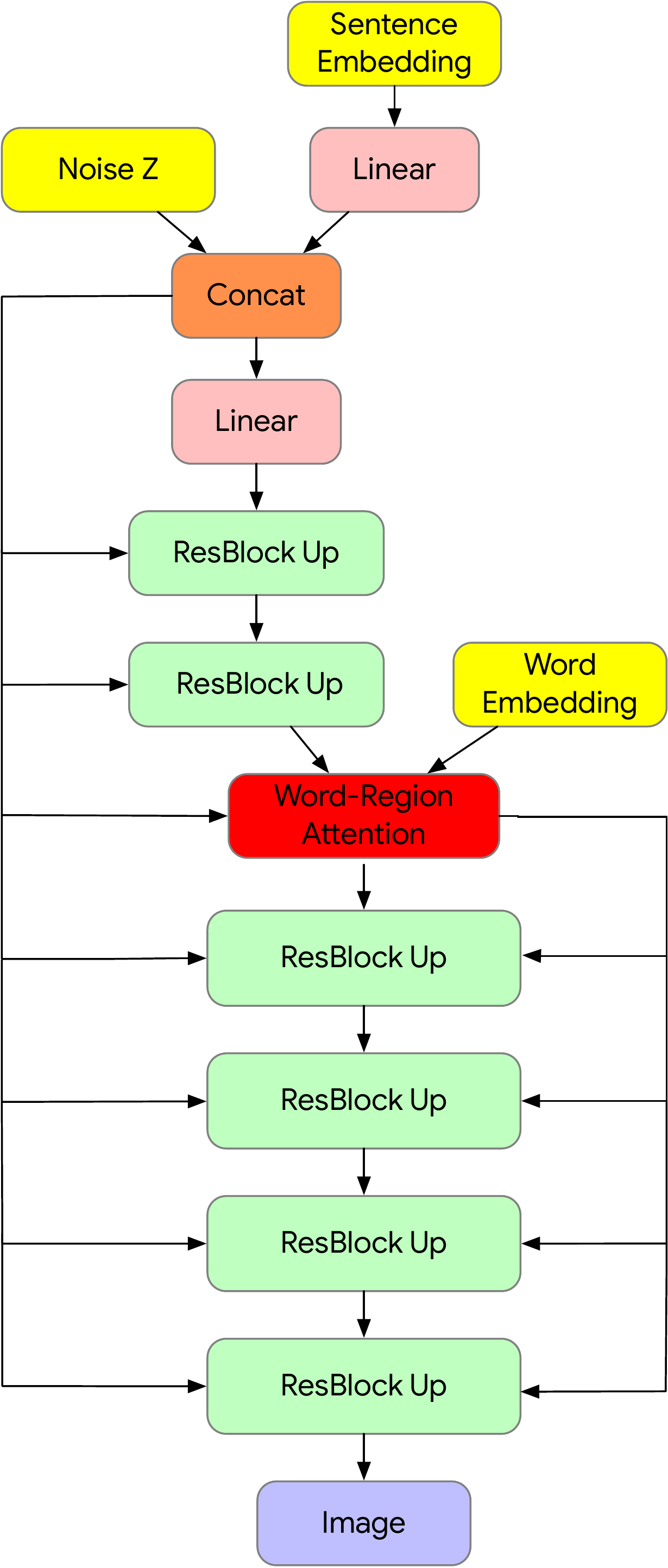}
        \caption{}
    \end{subfigure}%
    \hfill
    \begin{subfigure}[t]{0.33\textwidth}
        \centering
        \includegraphics[width=1.0\textwidth]{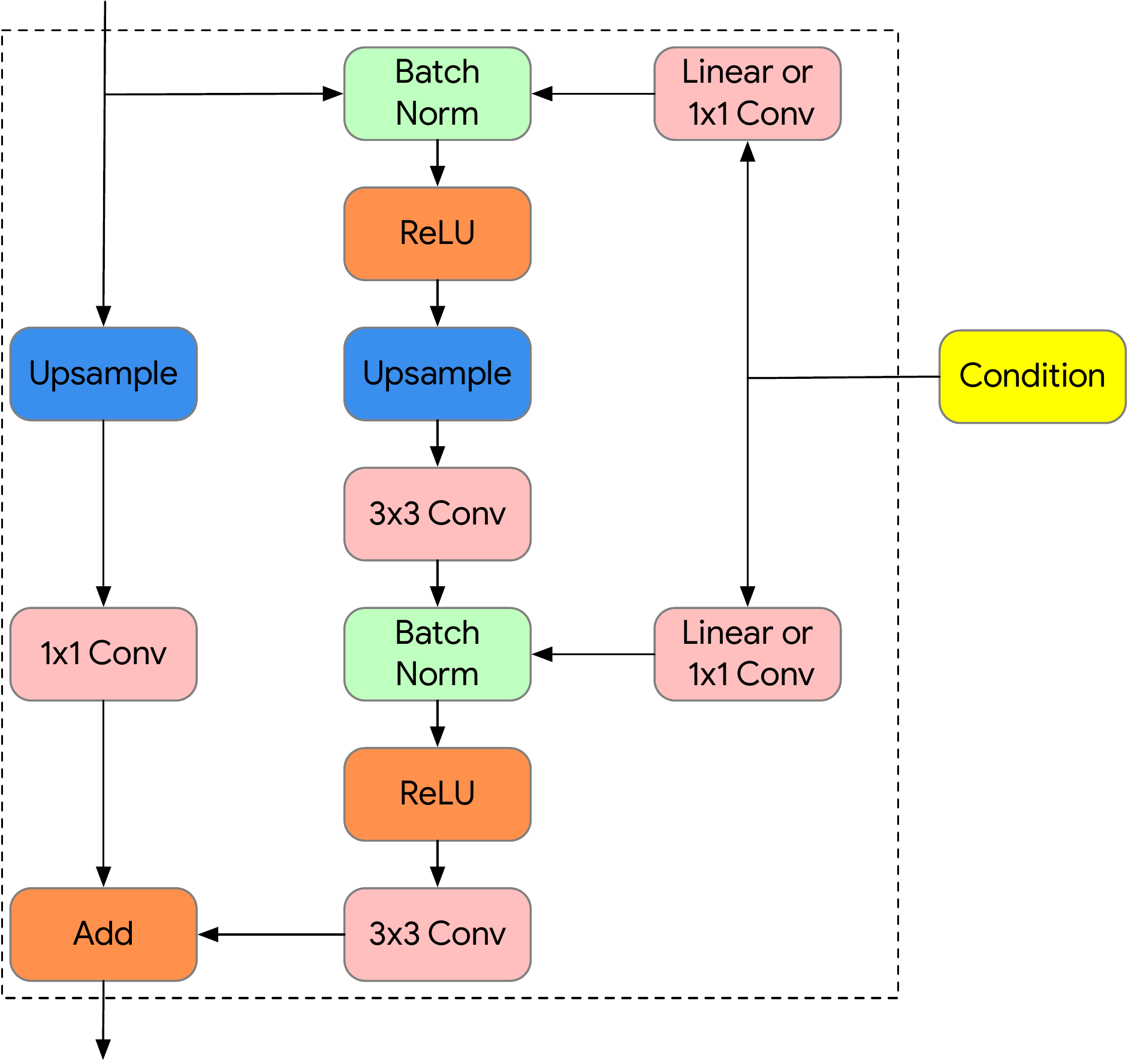}
        \caption{}
    \end{subfigure}%
    \hfill
    \begin{subfigure}[t]{0.2\textwidth}
        \centering
        \includegraphics[width=1.0\textwidth]{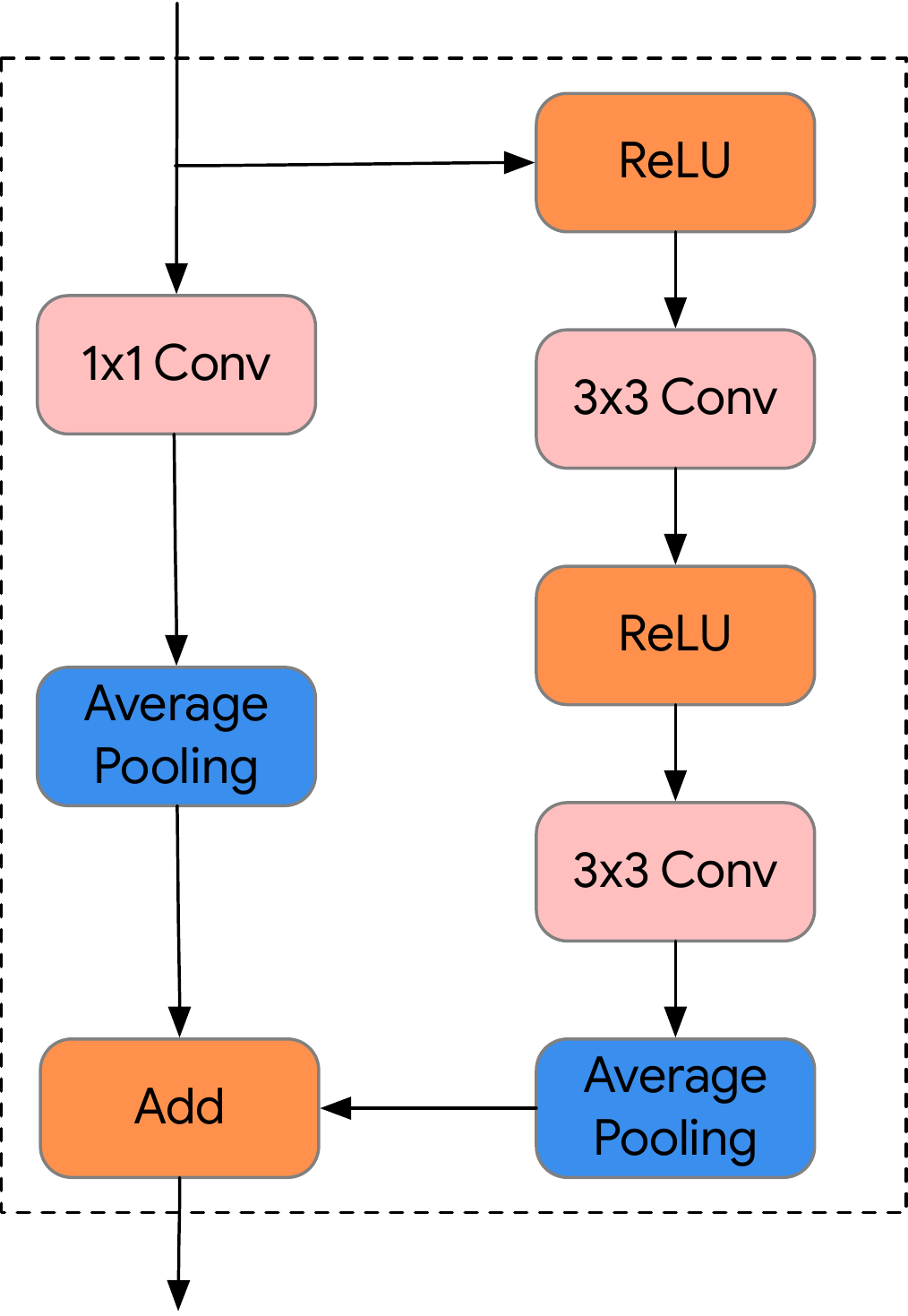}
        \caption{}
    \end{subfigure}
        \caption{(a) The generator achitecture for \CLGAN. (b) The residual block (ResBlock Up) of \CLGAN's generator. For the self-modulation ResBlock Up, the condition are noise $z$ and global sentence embedding. For attentional self-modulation ResBlock Up, the condition are noise $z$, global sentence embedding and attentional work context. 
        (c) The Residual Block (ResBlock Down) of \CLGAN's discriminator.}
        \label{fig:architecture_blocks}
\end{figure*}

Detailed generator and discriminator architectures can be found in Tables~\ref{tab:256_Generator} and \ref{tab:256_Discriminator} respectively. 
The details of the up-sampling block and down-sampling block are shown in Fig.~\ref{fig:architecture_blocks}.

\section{Human Evaluations} \label{appendix:human_evals}
\begin{figure*}[h]
    \centering
    \begin{subfigure}[t]{0.5\textwidth}
        \centering
        \includegraphics[width=1.0\textwidth]{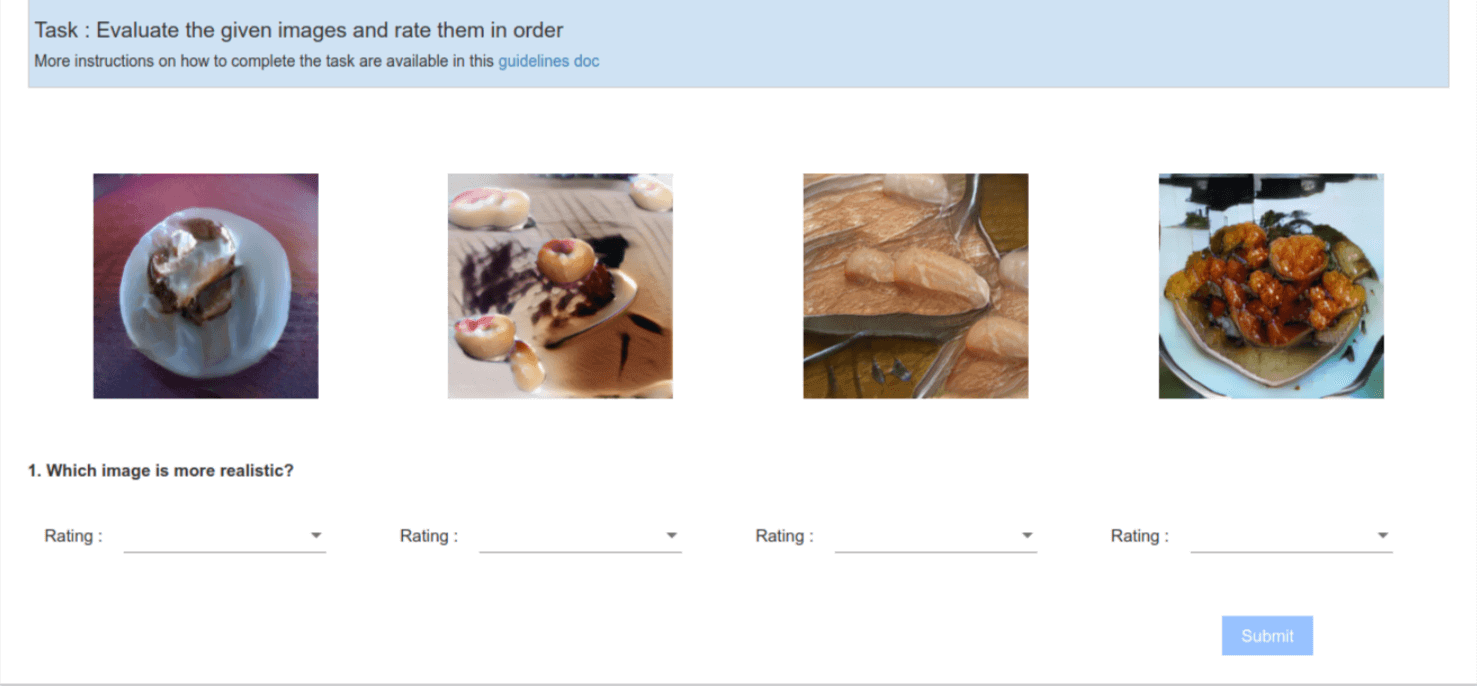}
        \caption{UI for ranking image realism.}
    \end{subfigure}%
    ~ 
    \begin{subfigure}[t]{0.5\textwidth}
        \centering
        \includegraphics[width=1.0\textwidth]{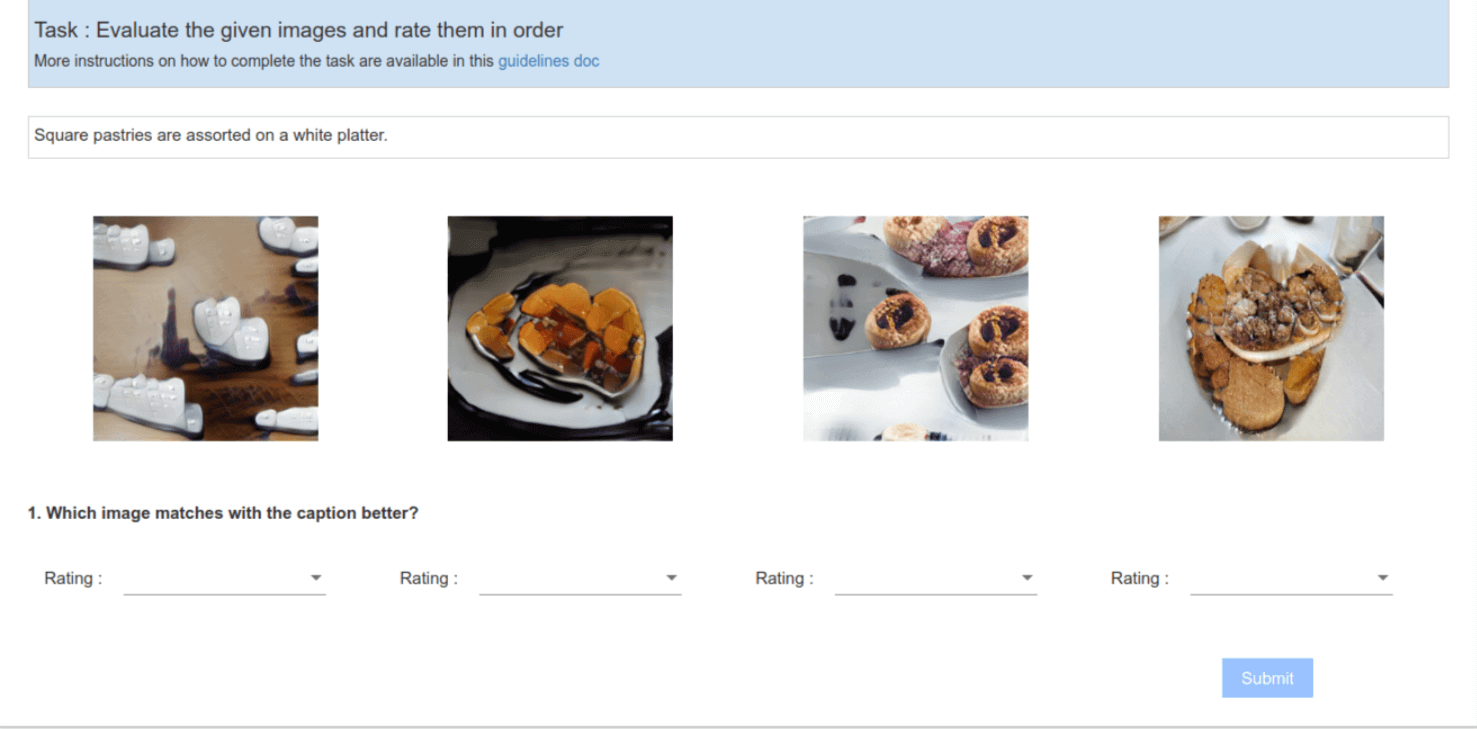}
        \caption{UI for ranking text alignment.}
    \end{subfigure}
    \caption{User interface for collecting human evaluations.}
    \label{fig:human_evals_ui}
\end{figure*}

The user interface shown to human evaluators is shown in Fig.~\ref{fig:human_evals_ui}. Users are requested to rank 4 images from best to worst on (1) image realism and (2) alignment to a given caption. The images are displayed in a random order.

\section{Similarities and differences between DAMSM and the proposed contrastive losses}
\begin{table}[h]
\begin{center}
\tiny
\resizebox{1.0\linewidth}{!}{%
\begin{tabular}{lccccc}
\hline
\textbf{Loss}  & \textbf{IS} $\uparrow$ & \textbf{FID} $\downarrow$  & \textbf{R-prec} $\uparrow$ & \textbf{SOA-C} $\uparrow$ & \textbf{SOA-I} $\uparrow$ \\
\hline
G & 23.69 & 34.70 & 40.44 & 21.61 & 38.13  \\
D & 25.81 & 26.63 & 56.62 & 28.58 & 49.36  \\
G + D (XMC-GAN) & \textbf{31.33} & \textbf{11.34} & \textbf{73.11} & \textbf{42.29} & \textbf{61.39}  \\
\hline
\end{tabular}
}
\end{center}
\vspace{-5pt}
\caption{Contrastive losses applied on the generator/discriminator.}
\label{tab:ablation_gd}
\vspace{-12pt}
\end{table}
\noindent
Our proposed contrastive losses bear several similarities to the DAMSM losses of AttnGAN. However, there are several key differences which are crucial to our strong performance: 
\begin{itemize}
    \item DAMSM losses are only used to train the \textit{generator} ($G$), while contrastive losses in XMC-GAN are designed to train the \textit{discriminator} ($D$) also. Features for contrastive losses are calculated from the different heads of the $D$ backbone. This allows $D$ to learn more robust and discriminative features, so XMC-GAN is less prone to mode collapse. This is a key reason that our model does not require multi-stage training. For training $G$, our contrastive losses are similar to DAMSM, which enforce consistency between generated images and conditional text descriptions. Table \ref{tab:ablation_gd} compares adding contrastive losses on $D$ and $G$ separately, which highlights the benefits of our proposed method of training the discriminator.
    \item Second, the motivation behind contrastive losses and DAMSM also differs. As described in Sec. 4.1, we propose maximizing the mutual information between intra-modality and inter-modality pairs. We do this by maximizing the lower bound through optimizing contrastive (InfoICE) losses, consistently using cosine distance as the similarity metric. In contrast, the DAMSM loss in AttnGAN is motivated by information retrieval. Their DAMSM module uses dot product in certain instances (Eq.~7 in AttnGAN), and  requires an additional normalization step (Eq.~8 in AttnGAN).
    \item Last, our training procedure is completely end-to-end, while AttnGAN needs a separate pretraining step. For AttnGAN, their DAMSM module undergoes a separate pretraining step before training the main generator / discriminator models.
\end{itemize}

\section{Qualitative Results} \label{appendix:qual}
\subsection{Effect of random noise on generated images}

In Sec.~6.1 of the main paper, we show that \CLGAN generated images are largely preferred by human raters. \CLGAN also significantly improves state-of-the-art FID scores. However, we also observe that the IS and SOA scores for CP-GAN are better than \CLGAN. We conjecture that the issue was with IS and SOA not penalizing intra-class mode dropping (\ie low diversity within a class or caption).

\begin{figure*}[h]
\small
\begin{center}
\begin{tabular}{>{\arraybackslash}m{0.08\linewidth}c@{\hspace{0.2em}}c@{\hspace{0.2em}}c@{\hspace{0.4em}}c@{\hspace{0.2em}}c@{\hspace{0.2em}}c@{\hspace{0.2em}}}
\multicolumn{1}{c}{\textbf{Caption}} & \multicolumn{3}{c}{\textbf{CP-GAN}} & \multicolumn{3}{c}{\textbf{\clgan}} \\ \cmidrule(l){2-4} \cmidrule(l){5-7}
 & \textbf{$z_1$} & \textbf{$z_2$} & \textbf{$z_3$} & \textbf{$z_1$} & \textbf{$z_2$} & \textbf{$z_3$} \\ \midrule

\compareexamplenoise{A newspaper wallpaper in a very modern bathroom.
}{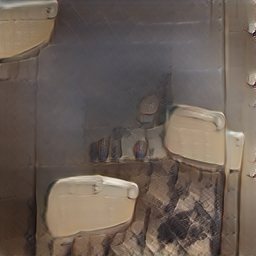} \\
\compareexamplenoise{A bus that is sitting in the street.}{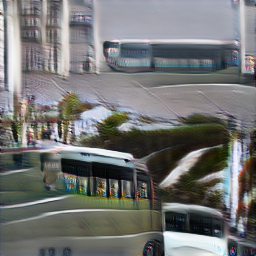} \\
\compareexamplenoise{an image of people outside playing frisbee}{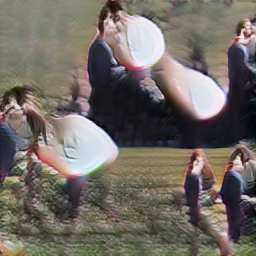} \\
\compareexamplenoise{A group of skiers are preparing to ski down a mountain.}{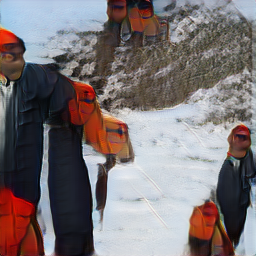} \\
\compareexamplenoise{A group of elephants walking in muddy water.}{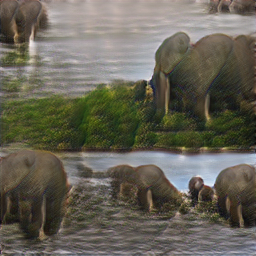}

\end{tabular}
\end{center}
\vspace{-10pt}
\caption{Comparison of CP-GAN and \clgan generated images for the same caption with different noise vectors.} \label{fig:compare_noise}
\end{figure*}

To verify this hypothesis, we conduct experiments to generate images from CP-GAN and \CLGAN conditioned on the same caption, but with varying noise vectors $z$. The comparison results are shown in Fig.~\ref{fig:compare_noise}. Both the captions and noise vectors used are selected at random. As shown in the figure, \CLGAN is able to generate diverse images (\eg, different view angles or compositions of the scene) for a fixed caption when different noise vectors are used. In contrast, CP-GAN generated images do not show much diversity despite conditioning on different noise vectors. This verifies our hypothesis that CP-GAN may have less diversity for the same class or caption. \CLGAN is able to generate high quality and diverse scenes even when conditioned on a single caption. 

\subsection{Effect of captions on generated images}
\begin{figure*}[h]
\small
\begin{center}
\begin{tabular}{>{\arraybackslash}m{0.11\linewidth}>{\arraybackslash}m{0.11\linewidth}>{\arraybackslash}m{0.11\linewidth}>{\arraybackslash}m{0.11\linewidth}>{\arraybackslash}m{0.11\linewidth}>{\arraybackslash}m{0.11\linewidth}>{\arraybackslash}m{0.13\linewidth}}
 \textbf{Real Image} & \textbf{Caption \#1} & \textbf{Caption \#2} & \textbf{Caption \#3} & \textbf{Caption \#4} & \textbf{Caption \#5} & \textbf{\cocoln} \\ \midrule

\compareexamplecaptions{The bus is pulling off to the side of the road.}{A bus pulls over to the curb close to an intersection.}{A bus driving in a city area with traffic signs.}{a public transit bus on a city street}{Bus coming down the street from the intersection}{in this image there are some vehicles on the road and behind the vehicles one big building is there on the right side there are some persons are walking on the street and the background is little bit sunny.}{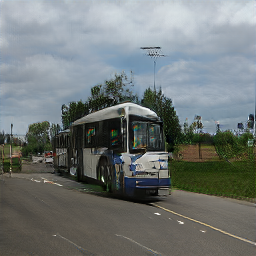}{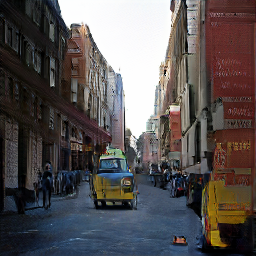}
\compareexamplecaptions{A group of people sitting around a table with laptops and notebooks.}{Seven people seated at table talking and working on computer devices.}{A group of people at a table working on small laptops.}{A group of people sitting at a table using computers.}{Several friends are visiting at a table with tablets.}{In the center of the image there is a table and there are people sitting around the table. We can see bottles, laptops and wires placed on the table. In the background there is a man standing. We can see a counter table, chairs and lights.}{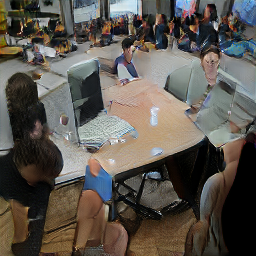}{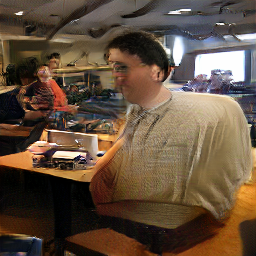}
\compareexamplecaptions{A group of people are walking and one is holding an umbrella.}{these people are walking together down a road}{Three young people walking behind a large crowd.}{Three men who are walking in the sand.}{A group of people walking down a road.}{In this image, in the middle there are some people walking, in the right side there is a man standing and he is holding a umbrella, in the background there are some cars, there is a bus, there are some green color trees, in the top there is a sky which is cloudy and in white color.}{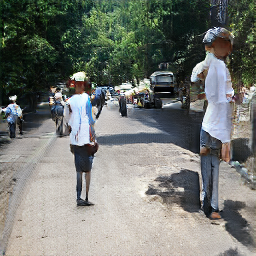}{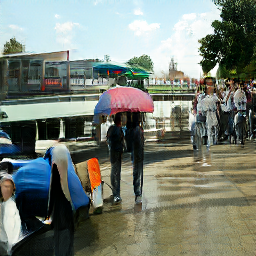}
\compareexamplecaptions{People are in a parking lot  beside the water, while a train is in the background.}{Colorful commuter train goes through a marina area on a cloudy day}{A  parking lot next to a marina next to a railroad}{Group of people standing beside their cars on a pier.}{A train crosses as a bunch of gathered vehicles watch.}{Bottom left side of the image there are two vehicles behind the vehicles there are few ships on the water and there are few people are standing. In the middle of the image there is a train on the bridge. Behind the train there are some trees and clouds. In the middle of the image there are two poles.}{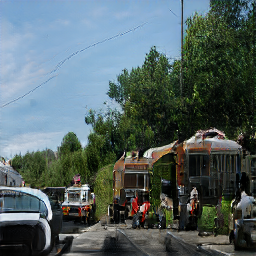}{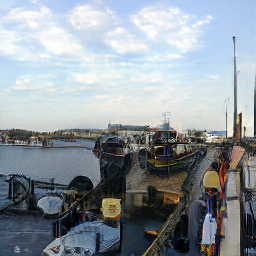}
\compareexamplecaptions{A calculator and cell phone lay on a desk in front of a keyboard}{A cell phone on top of a calculator near a computer keyboard.}{a table with a calculator and phone siting on it}{A picture of a cell phone Calculator and a computer.}{There is a phone on top of a calculator}{In the picture we can see a calculator which is black in color and on it there is a mobile phone and it is also black in color, in the background we can see a keyboard which is white in color placed on white paper on the wooden table.}{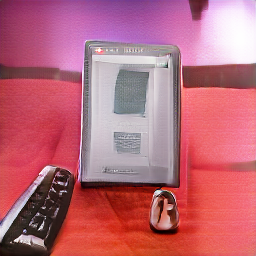}{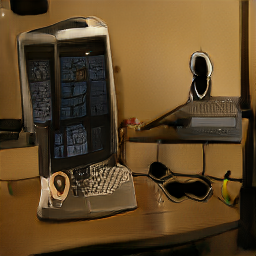}

\end{tabular}
\end{center}
\vspace{-10pt}
\caption{Generated images for varying captions from \cocoold and \cocoln corresponding to the same original image.} \label{fig:compare_captions}
\end{figure*}

In Fig.~\ref{fig:compare_captions}, we present several examples of \clgan generated images given different captions corresponding to the same original image.

\paragraph{Different \coco captions.} We observe that the generated images vary widely depending on the given caption, even if they are semantically similar. For example, we observe that in the first row, \clgan generated images for caption \#2 and caption \#3 produce very different images. For caption \#3, ``A bus driving in a city area with traffic signs.'', we observe that \clgan is able to generate features of a city, with high-rise buildings in the background, and a traffic light to the left of the image. In contrast, in caption \#2, which does not mention the city \clgan generates an image that shows the bus next to a curb, in agreement with the caption.

\paragraph{\coco compared to \cocoln captions.} We also observe distinct differences in generated images when conditioned on \coco as compared to \cocoln captions. \cocoln captions are much more detailed, which increases image generation difficulty. The increase in difficulty of \cocoln captions appears to lead to less coherent scenes in general as compared to the \coco model (\eg the third row of Fig.~\ref{fig:compare_captions}).

\subsection{Random samples}
\begin{figure*}[h]
\small
\begin{center}
\begin{tabular}{>{\arraybackslash}m{0.1\linewidth}@{\hskip 0.5em}c@{\hskip 0.1em}c@{\hskip 0.1em}c@{\hskip 0.1em}c@{\hskip 0.1em}>{\arraybackslash}m{0.1\linewidth}@{\hskip 0.5em}c@{\hskip 0.1em}c@{\hskip 0.1em}c@{\hskip 0.1em}c@{\hskip 0.1em}}
 \textbf{Caption} & \textbf{OP-GAN} & \textbf{SD-GAN} & \textbf{CP-GAN} & \textbf{\CLGAN} & \textbf{Caption} & \textbf{OP-GAN} & \textbf{SD-GAN} & \textbf{CP-GAN} & \textbf{\CLGAN} \\ \midrule

\compareexamplesmall{A woman holding a child looking at a cow.}{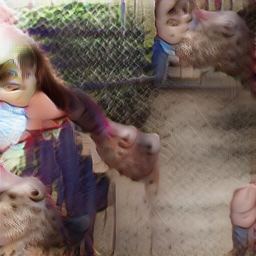} &
\compareexamplesmall{two brown dogs are laying next to each other}{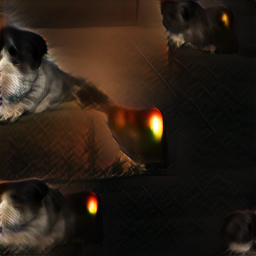} \\
\compareexamplesmall{A picture of a very tall stop sign.}{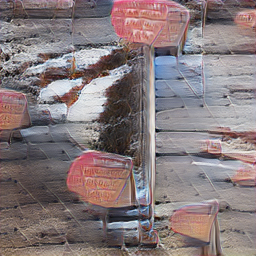} &
\compareexamplesmall{The boy hits the baseball with a bat.}{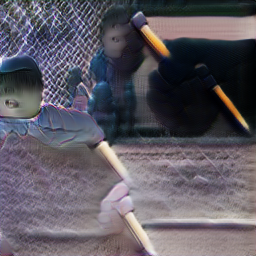} \\
\compareexamplesmall{A pelican near some boats that are docked.}{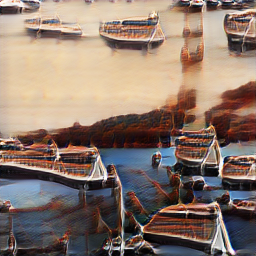} &
\compareexamplesmall{A picture of some food on a plate}{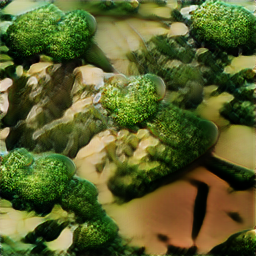} \\
\compareexamplesmall{A long boat is sitting on the clear water.}{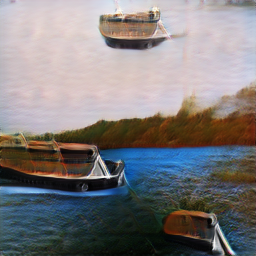} &
\compareexamplesmall{A woman throwing a frisbee with another person nearby}{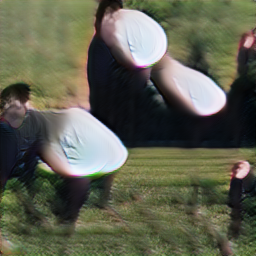} \\
\compareexamplesmall{A computer desk with a mouse and mouse pad.}{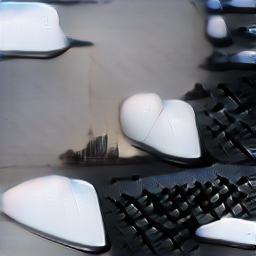} &
\compareexamplesmall{A bus that is sitting in the street.}{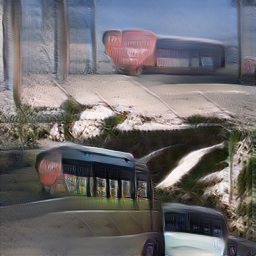} \\
\compareexamplesmall{a woman opening up a travel map}{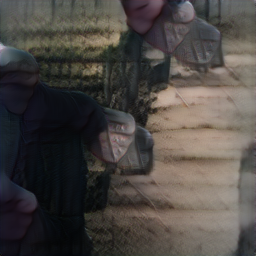} &
\compareexamplesmall{A tennis match in progress in an arena}{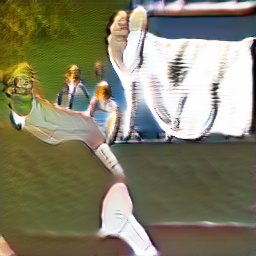} \\
\compareexamplesmall{A cat sitting beside a bunch of bananas.}{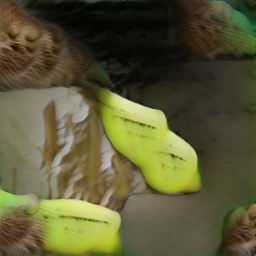} &
\compareexamplesmall{Two geese walking in a parking lot.}{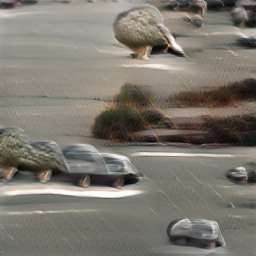} \\
\compareexamplesmall{A water hydrant on the sidewalk with plants nearby}{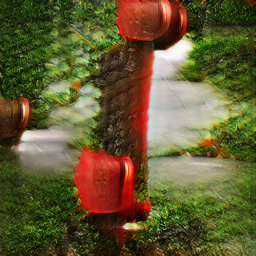} &
\compareexamplesmall{A parade in historical clothing is walking down the street.}{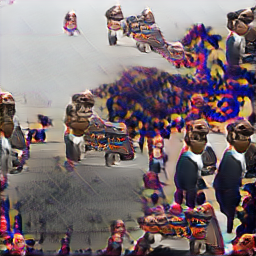} \\
\compareexamplesmall{London transportation with no passengers sitting on the street.}{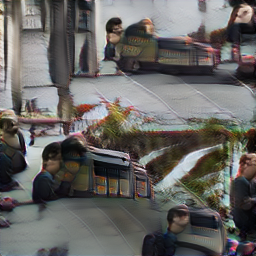} &
\compareexamplesmall{Woman showing delight with plated chocolate desert dish .}{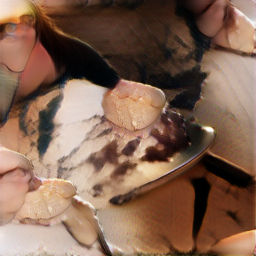} \\
\compareexamplesmall{A desk containing a black laptop, candy, money, and several bananas.}{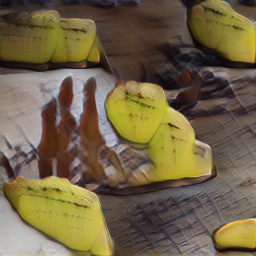} &
\compareexamplesmall{A train traveling down a track in the country.}{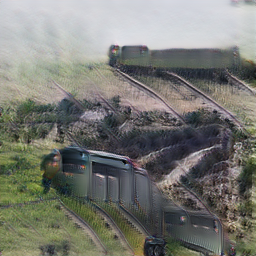} \\
\compareexamplesmall{A girl reading a book in bed with a cat}{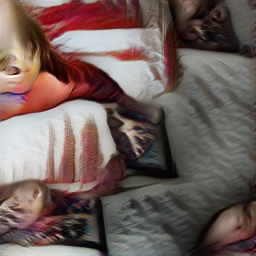} &
\compareexamplesmall{A bedroom scene with focus on the bed.}{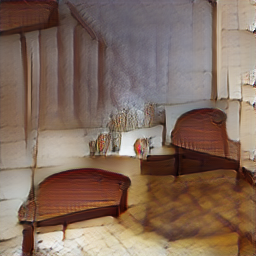} \\
\compareexamplesmall{A boat in the middle of the ocean.}{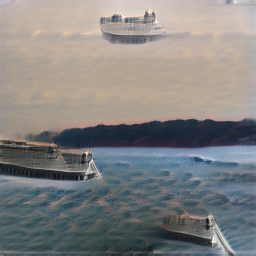} &
\compareexamplesmall{A plate of breakfast food including eggs and sausage.}{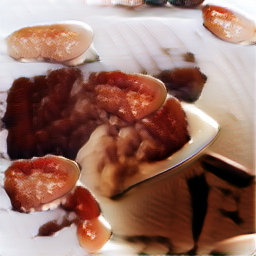} \\

\end{tabular}
\end{center}
\vspace{-10pt}
\caption{Generated images for random examples from \cocoold.} \label{fig:compare_coco_random}
\end{figure*}

\paragraph{\cocoold} Random qualitative samples from \cocoold are presented in Fig.~\ref{fig:compare_coco_random}. We observe that even over randomly selected captions, \clgan appears to generate images that are significantly clearer and more coherent. Scenes often depict clear objects, as compared to previous methods.

\begin{figure*}[h]
\small
\begin{center}
\begin{tabular}{>{\arraybackslash}m{0.25\linewidth}@{\hskip 0.5em}c@{\hskip 0.1em}c@{\hskip 0.1em}c@{\hskip 0.1em}c@{\hskip 0.1em}}
\textbf{Caption} & \textbf{Original} & \textbf{AttnGAN} & \textbf{\textsc{TReCS}} & \textbf{\CLGAN} \\ \midrule  \addlinespace[0em]

\compareexampleln{In this picture we can see a pole in front, in the bottom there are some leaves, in the background we can see a white color and black color cars, on the right side of the image we can see a tree, in the background there is a building and a hill.}{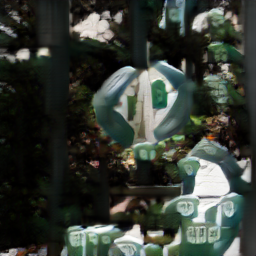}{000000458702} \\
\compareexampleln{In this image we can see zebra and giraffe standing in grass, And there are so many plants, lake with water, mountain with trees.}{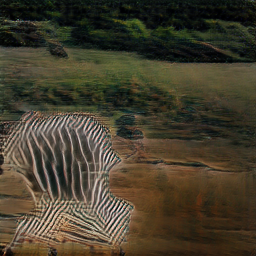}{000000159977} \\
\compareexampleln{In this image we can see both of the children are standing, and smiling and cooking, in front here is the stove and pan on it, here is the spoon, at side here is the vessel, and at back here is the table, here is the wall, and here is the glass door.}{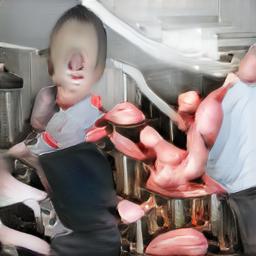}{000000512929} \\
\compareexampleln{In this image there are group of persons who are sitting around the table in a restaurant and having some food and there are water glasses on the table,at the background of the image there is a door,mirror and some paintings attached to the wall.}{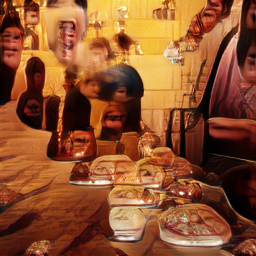}{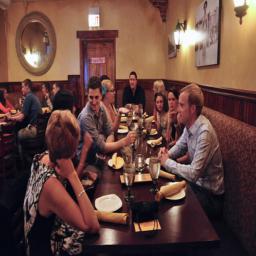} \\
\compareexampleln{Here we can see a woman sitting in the briefcase. And this is wall.}{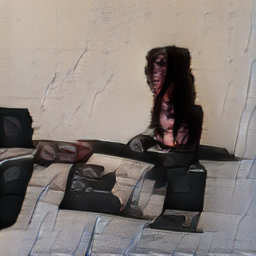}{000000127660} \\
\compareexampleln{There is a man in white color shirt, wearing a black color tie, standing. In the background, there is a yellow wall near the white ceiling.}{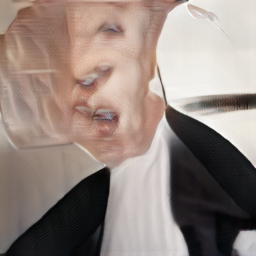}{000000476415} \\
\compareexampleln{The picture consists of food items on a white color plate like object.}{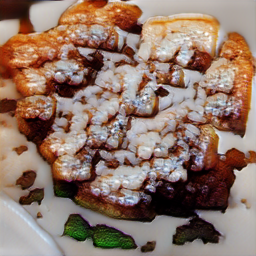}{000000015278} \\
\compareexampleln{In this image i can see person holding a bat and a wearing a white helmet. He is wearing blue shirt and white pant. At the back side I can see three person sitting. There is a net. The person is holding a umbrella which is in green and white color. Back Side i can see vehicle.}{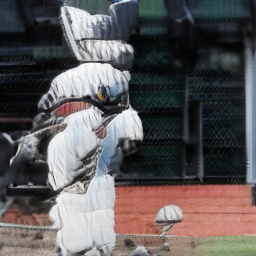}{000000357816} \\
\compareexampleln{Here we can see a bench and this is road. There are plants and this is grass. In the background there is a wall.}{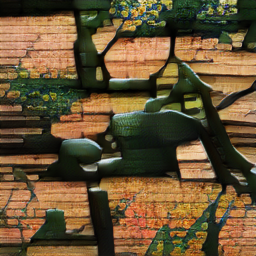}{000000461573} \\
\compareexampleln{This image consists of refrigerator. On that there are cans and boxes. There is light on the top. There is magnum sticker on refrigerator.}{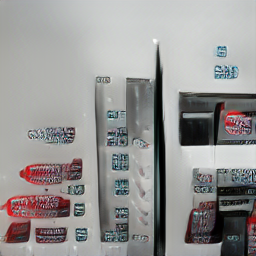}{000000554838} \\


\end{tabular}
\end{center}
\vspace{-10pt}
\caption{Original and generated images for random examples from \cocoln.} \label{fig:compare_ln_coco}
\end{figure*}

\paragraph{\cocoln} Random qualitative samples from \cocoln are presented in Fig.~\ref{fig:compare_ln_coco}. The longer captions increase the challenge of realistic text-to-image synthesis, but we observe clear improvements from previous methods in most images. In particular, \clgan appears to generate objects and people that are more clear and distinct.

\begin{figure*}[h]
\small
\begin{center}
\begin{tabular}{>{\arraybackslash}m{0.2\linewidth}@{\hskip 0.5em}c@{\hskip 0.1em}c@{\hskip 0.1em}>{\arraybackslash}m{0.2\linewidth}@{\hskip 0.5em}c@{\hskip 0.1em}c@{\hskip 0.1em}}
\textbf{Caption} & \textbf{Original} & \textbf{\CLGAN} & \textbf{Caption} & \textbf{Original} & \textbf{\CLGAN} \\ \midrule  \addlinespace[0em]

\compareexamplelnoi{In this picture I can see the cars on the grass  in the top right hand side there is a vehicle. In the background there may be the buildings.}{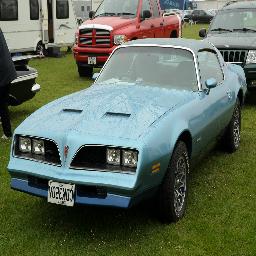} &
\compareexamplelnoi{In this image I can see a mirror with some text written on it. In the background  I can see a car  the trees and the buildings with some text written on it.}{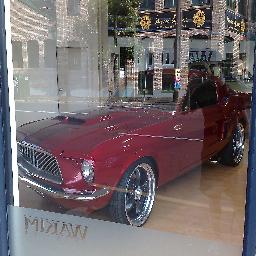} \\
\compareexamplelnoi{In this image  I can see a cat on a sidewalk and I can see a dark color.}{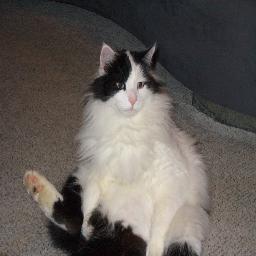} &
\compareexamplelnoi{In this image we can see people sitting on chairs. Also we can see packets on chairs. There are two people standing. Also we can see cupboards with books. And there is a pillar. And there is a table ...}{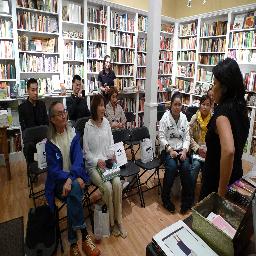} \\
\compareexamplelnoi{In this image  we can see vehicles  a fence and a pole. At the top  there is sky. At the bottom  there are plants and we can see grass.}{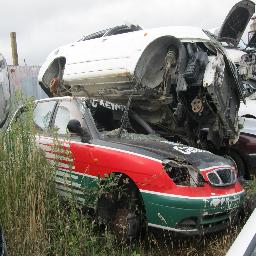} &
\compareexamplelnoi{In this picture we can see a grill meat piece in black plate which is placed on the wooden table top.}{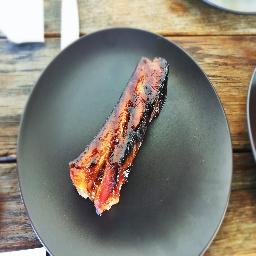} \\
\compareexamplelnoi{In front of the image there is a person running on the track. Beside the track there is a sponsor board. At the bottom of the image there is grass on the surface.}{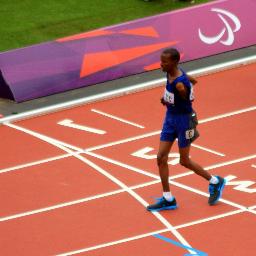} &
\compareexamplelnoi{In this image in the foreground we can see a sculpture and in the background we can see many branches of a tree.}{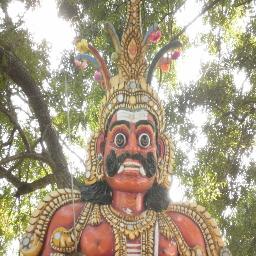} \\
\compareexamplelnoi{This is an aerial view and here we can see buildings and trees. At the top  there is sky.}{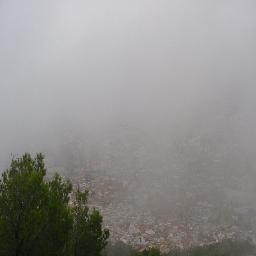} &
\compareexamplelnoi{In front of the image there is an army personnel holding some objects in his hand. Behind the person there are a few army personnel. In the background of the image there are photo frames and doors on ...}{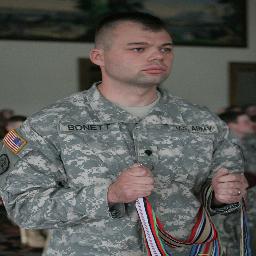} \\
\compareexamplelnoi{In this image I can see cake on the table. There is hand of a person holding the knife  also there are hands of another person holding food item in one hand. And there are some other objects.}{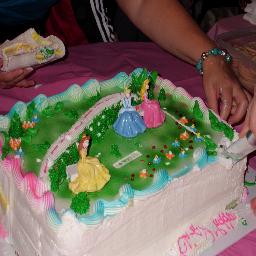} &
\compareexamplelnoi{In this picture  we see a plastic glass containing the ice cream is placed on the white table. We see the tissue papers and a paper glass are placed on the table. In the background  we see a grey color object is placed ...}{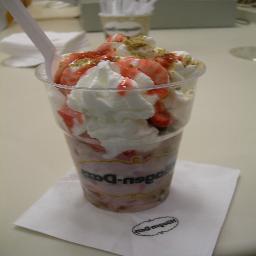} \\
\compareexamplelnoi{In this image we can see a bunch of flowers to the plants. We can also see the wooden surface.}{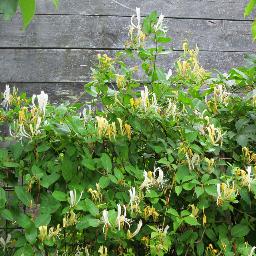} &
\compareexamplelnoi{In this picture I can see few plants with leaves and I can see the flowers.}{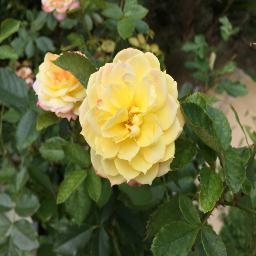} \\
\compareexamplelnoi{In the foreground  I can see grass  a fence  a net  light poles and wires. In the background  I can see water  house  plants  some objects  the trees and the sky.}{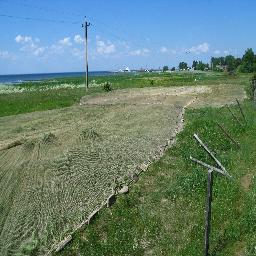} &
\compareexamplelnoi{It is an edited image with different shaped designs.}{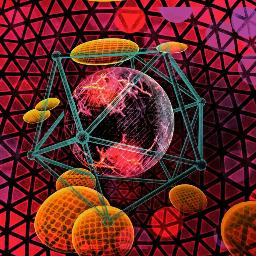} \\
\compareexamplelnoi{In this image  there is dried grass on the ground. In the top left side of the image  I can see a tree. In the background there is sky.}{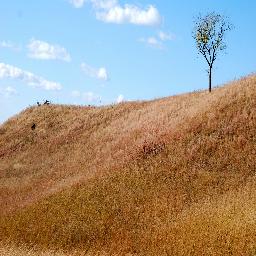} &
\compareexamplelnoi{In this image  there are birds on a pathway and I can see a duck in the water.}{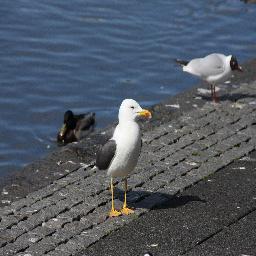} \\
\compareexamplelnoi{In this image I can see a pen which is black in color on the white colored surface.}{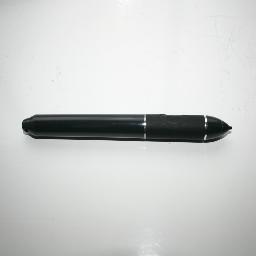} &
\compareexamplelnoi{In this image I can see the cat on the mat and I can see few objects.}{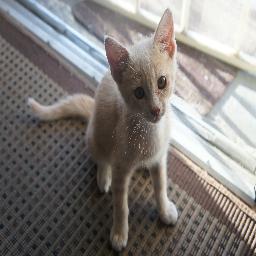} \\


\end{tabular}
\end{center}
\vspace{-10pt}
\caption{Original and generated images for random examples from \lnoi.} \label{fig:compare_ln_oi}
\end{figure*}

\paragraph{\lnoi} Random qualitative samples from \lnoi are presented in Fig.~\ref{fig:compare_ln_oi}. As this dataset was previously untested on, we simply display the original images against \clgan generated images. Despite the increase in complexity and diversity of images, \clgan generates very strong results, with especially convincing scene generation capability (\eg first column, second and third last rows). We hope that our results will inspire future work to advance on tackling this very challenging dataset.


\begin{table*}[tbh]
\begin{subtable}[t]{0.49\textwidth}
\centering
\resizebox{\linewidth}{!}{%
\begin{tabular}{l}
\hline
\hline
$z \in \mathbb{R}^{128} \sim  \mathcal{N}(0, I)$,
$e_{s} \in \mathbb{R}^{768}$, $e_w \in \mathbb{R}^{T \times 768}$ \\
\hline
Linear ($768$) $\rightarrow 128\qquad$ \# projection for $e_s$ \\
\hline
Linear ($128+128$) $\rightarrow 4 \times 4 \times 16ch$  \\
\hline
Self-modulation ResBlock up  $\rightarrow 8 \times 8 \times  16ch$\\ 
\hline
Self-modulation ResBlock up  $\rightarrow 16 \times 16 \times  8ch$ \\ 
\hline
Linear Layer ($8ch$) $\rightarrow 768\qquad$ \# projection for attention\\
\hline
Attentional Self-modulation ResBlock up  $\rightarrow 32 \times 32 \times  8ch$ \\
\hline
Attentional Self-modulation ResBlock up  $\rightarrow 64 \times 64 \times  4ch$ \\
\hline
Attentional Self-modulation ResBlock up  $\rightarrow 128 \times 128 \times  2ch$ \\
\hline
Attentional Self-modulation ResBlock up  $\rightarrow 256 \times 256 \times  ch$ \\
\hline
Attentional Self-modulation, $3\times3$ Conv $\rightarrow 256 \times 256 \times  3$ \\
\hline
\hline
\end{tabular}
}
\caption{Generator} 
\label{tab:256_Generator}
\end{subtable}
\hspace{\fill}
\begin{subtable}[t]{0.49\textwidth}
\centering
\resizebox{\linewidth}{!}{%
\begin{tabular}{l}
\hline
\hline
RGB images $x \in \mathbb{R}^{256\times256\times3}$,
$e_{s} \in \mathbb{R}^{768}$, $e_w \in \mathbb{R}^{T \times 768}$ \\
\hline
ResBlock down  $\rightarrow 128 \times 128 \times  ch$\\ 
\hline
ResBlock down  $\rightarrow 64 \times 64 \times  2ch$\\ 
\hline
ResBlock down  $\rightarrow 32 \times 32 \times  4ch$\\ 
\hline
ResBlock down  $\rightarrow 16 \times 16 \times  8ch$\\ 
\hline
Linear ($4ch$)  $\rightarrow 768 \qquad$ \# projection for word-region contrastive\\ 
\hline
ResBlock down  $\rightarrow 8 \times 8 \times  8ch$\\ 
\hline
ResBlock down  $\rightarrow 4 \times 4 \times  16ch$\\ 
\hline
ResBlock  $\rightarrow 4 \times 4 \times  16ch$\\ 
\hline
Global sum pooling\\
\hline
Linear ($768$)  $\rightarrow 16ch\qquad$ \# projected($e_s) \cdot h$\\
\hline
Linear ($16ch$)  $\rightarrow 1$ \\
\hline
\hline
\end{tabular}
}
\caption{Discriminator}
\label{tab:256_Discriminator}
\end{subtable}
\caption{\CLGAN generator and discriminator architectures.} 
\label{tab:256_architecture}
\end{table*}


\end{document}